\documentclass{article}

    \PassOptionsToPackage{numbers}{natbib}

 \usepackage[main, final]{neurips_2026}

\usepackage[utf8]{inputenc} 
\usepackage[T1]{fontenc}    
\usepackage{hyperref}       
\usepackage{url}            
\usepackage{booktabs}       
\usepackage{amsfonts}       
\usepackage{nicefrac}       
\usepackage{microtype}      
\usepackage[table]{xcolor}         
\usepackage{graphicx}
\usepackage{multirow}
\usepackage{rotating} 
\usepackage{wrapfig}
\usepackage{amsmath}
\usepackage{cleveref}
\usepackage{tabularx}
\usepackage{pifont}
\newcommand{\posdiff}[1]{\scriptsize\color{red}{(+#1)}} 
\newcommand{\negdiff}[1]{\scriptsize\color{blue}{(-#1)}} 
\definecolor{gcolor1}{RGB}{255,240,220} 
\definecolor{gcolor2}{RGB}{230,240,255}

\title{Facts Don't Speak Louder than Words: \\ The Behavioral Essence of Long-CoT Distillation}

\author{
 \textbf{Yongcan Yu\textsuperscript{1,2}\thanks{Work done during an internship at Meituan}},
 \textbf{Jian Liang\textsuperscript{1,2}\thanks{Corresponding author}},
  \textbf{Lingxiao He\textsuperscript{3}},
 \textbf{Kuangpu Guo\textsuperscript{1,4}},
\\
\textbf{Yanbo Wang\textsuperscript{1,2}},
\textbf{Shuo Lu\textsuperscript{1,2}},
 \textbf{Meng Wang\textsuperscript{3}},
 \textbf{Ran He\textsuperscript{1,2}}
\\
 \textsuperscript{1}NLPR \& MAIS, Institute of Automation, Chinese Academy of Sciences\\
 \textsuperscript{2}School of Artificial Intelligence, University of Chinese Academy of Sciences\\
 \textsuperscript{3}Meituan
 \textsuperscript{4}University of Science and Technology of China\\
\texttt{\{yuyongcan0223, liangjian92\}@gmail.com}
}

\begin{document}

\maketitle

\begin{abstract}

While long chain-of-thought (CoT) distillation significantly improves the reasoning abilities of language models, its underlying mechanism remains poorly understood.
Current pipelines generally rely on rejection sampling to gather reasoning traces across diverse prompts, assuming that factual correctness and broad problem coverage are indispensable.
Surprisingly, we find that training exclusively on flawed reasoning trajectories or a restricted prompt set yields nearly comparable performance.
Driven by this counterintuitive finding, we propose the \textit{Behavioral Hypothesis}: \textbf{the essence of Long-CoT distillation is behavioral alignment with structured reasoning patterns, rather than factual knowledge memorization.}
We then validate this by demonstrating substantial reasoning gains even when models are fine-tuned exclusively on previously mastered problems—effectively isolating behavioral alignment from novel knowledge injection.
Having established that reasoning patterns are the true bottleneck, we investigate how to learn them across varying levels of complexity optimally.
We reveal a critical interplay: while the recent probability-based loss excels at learning simple patterns, the standard cross-entropy loss is essential for scaling to complex CoTs.
We believe these insights demystify CoT distillation and provide a principled foundation for the future development of reasoning models.
Code is available on \href{https://github.com/yuyongcan/Long-CoT-Distillation}{https://github.com/yuyongcan/Long-CoT-Distillation}.
\end{abstract}

\section{Introduction}
Large reasoning models~\citep{guo2025deepseek,jaech2024openai,sun2025survey,wang2025comprehensive} have recently garnered widespread attention due to their remarkable performance in handling complex tasks (e.g., mathematics~\citep{cobbe2021training,hendrycks2measuring,lu2026mllms,yu2026understanding}, coding~\citep{jimenez2024swe,jain2025livecodebench}, scientific reasoning~\citep{rein2024gpqa,phan2025humanity}, and others~\citep{lu2025deepresearch,lu2026one}).
Consequently, Long Chain-of-Thought (Long-CoT) distillation has emerged as a widely adopted post-training paradigm to endow models with these advanced reasoning capabilities~\citep{luo2025deconstructing,zhang2025thyme}.
However, while Long-CoT distillation significantly improves the reasoning abilities of language models, the underlying mechanism remains poorly understood.

Current distillation pipelines heavily depend on rejection sampling~\citep{tong2024dart} to curate correct reasoning traces across diverse prompts~\citep{luo2025deconstructing,zhang2025thyme,guan2025rstar}.
This practice operates under the implicit assumption that CoT correctness and broad problem coverage are highly important for effective distillation.
In this work, we revisit these assumptions across model families (Qwen3, Qwen2.5, and Llama 3.1) and two training losses.
First, we observe that models fine-tuned exclusively on incorrect reasoning trajectories achieve performance nearly comparable (\textbf{97.7\%}) to those trained on correct samples.
In these settings, localized factual errors do not erase the value of otherwise structured reasoning demonstrations.
Furthermore, we show that in data-constrained scenarios, mixing incorrect trajectories into the training set acts as an effective strategy to reinforce the model's behavioral alignment with step-by-step reasoning.
Second, we explore the role of prompt diversity.
Under a fixed trajectory budget, we observe that heavily sampling from a highly restricted set of prompts (e.g., 500) yields performance virtually identical to training on a vastly diverse dataset.
Thus, broad problem coverage is not the dominant bottleneck under a fixed trajectory budget.

Driven by these observations, we propose the \textit{Behavioral Hypothesis}: learning structured reasoning behaviors is a central driver of Long-CoT distillation, beyond memorization of perfectly correct and diverse factual knowledge.
Models learn to organize pre-existing knowledge into sustained, step-by-step reasoning.
Two complementary observations support this account: models trained on incorrect trajectories improve even on the training prompts, and models fine-tuned on problems they can already solve still improve on held-out benchmarks.
The latter comparison reduces the role of novel problem-specific knowledge without assuming that knowledge transfer is absent altogether.
Additional science-domain training, alternative teachers, newer benchmarks, and representation comparisons provide further evidence in \Cref{app:rebuttal}.

Finally, building upon the importance of reasoning patterns, we investigate how CoT complexity interacts with the choice of optimization objective. While recent research broadly advocates replacing the standard cross-entropy loss ($-\log p$) with probability-based loss ($-p$) to enhance reasoning acquisition~\citep{wu2026on,zhu2026anchored,li2025beyond}, our experiments across progressively complex datasets reveal a critical divergence.
Specifically, although the $-p$ loss yields notable gains over $-\log p$ on simpler CoTs, it suffers from pronounced performance degradation as reasoning complexity increases. Conversely, $-\log p$ loss scales robustly, effectively capturing the benefits of complex, lengthy reasoning traces.
These findings highlight the absence of a universally superior loss function and underscore the necessity of tailoring the optimization objective to the intrinsic complexity of the data.
In summary, our main contributions are as follows:
\begin{itemize}

\item \textit{Challenging Prevailing Distillation Assumptions:} We empirically deconstruct the deeply rooted assumptions that factual correctness and broad problem coverage are indispensable for effective Long-CoT distillation.
    
    \item \textit{Proposing and Empirically Supporting the Behavioral Hypothesis:} We introduce the Behavioral Hypothesis, providing strong empirical evidence that the essence of reasoning distillation lies in structural behavioral alignment (internalizing step-by-step methodologies) rather than factual knowledge memorization.
    
    \item \textit{Investigating Complexity-Aware Optimization:} We uncover a critical divergence in learning dynamics, demonstrating that the efficacy of optimization objectives (e.g., $-p$ vs. $-\log p$) is fundamentally dictated by the intrinsic complexity of the reasoning trajectories.
\end{itemize}

By illuminating the fundamental mechanisms of reasoning distillation, we believe our study provides actionable guidance for data curation and objective optimization in Long-CoT distillation, paving the way for more efficient and robust reasoning models.

\section{Experimental Setup}

Unless otherwise specified, all experiments in this paper follow the default setup described in this section. For experiments that require different data construction, sampling strategies, or training hyperparameters, we explicitly describe the deviations in the corresponding sections.

\subsection{Training and Evaluation Details}

\paragraph{Models and training configuration.}
We conduct experiments on a diverse set of open-source language models, including Qwen3-1.7B-Base, Qwen3-4B-Base, Qwen3-8B-Base, Qwen3-14B-Base~\citep{yang2025qwen3}, Qwen2.5-7B~\citep{yang2024qwen2}, and Llama-3.1-8B-Instruct~\citep{grattafiori2024llama}. Unless otherwise noted, our default supervised fine-tuning (SFT) dataset strictly consists of correct CoT trajectories.
These are constructed by sampling responses from Qwen3-4B~\citep{yang2025qwen3} on the DAPO-Math-17K~\citep{yu2025dapo} prompt set and applying rigorous rejection sampling~\citep{tong2024dart} to filter out any reasoning traces leading to incorrect answers.
We perform full-parameter fine-tuning using \texttt{verl}~\citep{sheng2025hybridflow}. Unless specified otherwise, we train for 3 epochs with global batch size 128 and learning rate $10^{-5}$. Both objectives use AdamW ($\beta_1=0.9$, $\beta_2=0.95$, weight decay 0.01), cosine decay with 10\% warmup, gradient clipping at 1.0, and bf16 precision; we do not tune these settings separately for the two losses. Simple-CoT training uses 1 epoch, and the micro-batch size varies with model/context length (typically 4 per GPU, 2 for the 8B easy-set comparison, and 1 for 32K contexts).

\paragraph{Evaluation benchmarks and metrics.}
We evaluate the fine-tuned models on mathematical reasoning benchmarks and broader reasoning benchmarks. For mathematical reasoning, we use 5 widely-used benchmarks, including AIME 2024~\citep{aim2024aime}, AMC23~\citep{maa2023amc}, Minerva Math~\citep{lewkowycz2022solving}, Olympiad Bench~\citep{he2024olympiadbench}, and MATH~\citep{hendrycks2measuring}. For MATH, we evaluate on a sampled subset of 2,000 examples. Following the evaluation protocol of Qwen2.5-Math~\citep{yang2024qwen2math}, we sample 16 responses with a temperature of 1.0 for each test problem. To rigorously assess the model's inherent reasoning generation capability, we report the \textit{avg@k} (averaged \textit{pass@1}) across all sampled responses.
For broader reasoning ability, we evaluate on GPQA-Diamond~\citep{rein2024gpqa} for science reasoning and MMLU-Pro~\citep{wang2024mmlu} for general reasoning, and LiveCodeBench-v2~\citep{jain2025livecodebench} for coding. These evaluations are conducted using the Evalchemy framework~\citep{Evalchemy} with a temperature of 0.6. For GPQA-Diamond, we sample 6 responses per example, while for MMLU-Pro and LiveCodeBench-v2, we sample 3 responses per example. Similar to the mathematical benchmarks.
All our experiments were conducted on 8 NVIDIA A100-80G GPUs.

\subsection{Training Objectives}

We conduct experiments with two widely-used SFT losses: the standard cross-entropy loss (\(-\log p\)) and the recently popularized plain probability-based loss (\(-p\)).

\paragraph{Standard supervised fine-tuning.}
Given a training dataset \(\mathcal{T}\) consisting of input-output pairs \((x, \tilde{y})\), where \(\tilde{y} = (y_1, \ldots, y_N)\) denotes the target reasoning trajectory and final answer, the model is trained by modeling token-level conditional probabilities \(p_\theta(y_t \mid y_{<t}, x)\). The standard SFT objective minimizes the negative log-likelihood (NLL), which is equivalent to the cross-entropy loss:
\begin{equation}
    \mathcal{L}_{\log p}(\theta)
    =
    \mathbb{E}_{(x,\tilde{y}) \sim \mathcal{T}}
    \left[
    \sum_{t=1}^{N}
    -\log p_\theta(y_t \mid y_{<t}, x)
    \right].
\end{equation}

\paragraph{SFT with plain-\(p\) objective.}
Since standard NLL optimization can sometimes lead to overfitting and poor generalization in lengthy generation tasks, recent works~\citep{wu2026on,zhu2026anchored,li2025beyond} have proposed replacing it with a probability-based objective. This formulation directly maximizes the token probabilities:
\begin{equation}
    \mathcal{L}_{p}(\theta)
    =
    \mathbb{E}_{(x,\tilde{y}) \sim \mathcal{T}}
    \left[
    \sum_{t=1}^{N}
    - p_\theta(y_t \mid y_{<t}, x)
    \right].
\end{equation}



\section{Rethinking the Role of Correctness and Diversity in Distillation}
\label{sec:findings}
Long-CoT distillation is commonly built upon two intuitive assumptions: the distilled trajectories should be factually correct, and the training prompts should be sufficiently diverse~\citep{guo2025deepseek,luo2025deconstructing,zhang2025thyme,guan2025rstar}.
These assumptions naturally motivate the standard practice of using rejection sampling~\citep{tong2024dart} to curate correct reasoning traces across a broad set of problems.

In this section, we systematically revisit both assumptions. We first investigate whether the factual correctness of CoT trajectories is strictly essential by comparing models fine-tuned on correct versus incorrect reasoning traces under two optimization objectives.
We then examine whether incorrect trajectories can still provide meaningful supervision when correct samples are scarce.
Finally, we control the total volume of training trajectories while varying the number of unique prompts to determine if prompt diversity itself is a key driver of reasoning improvement. Across these controlled settings, we consistently find that neither CoT correctness nor prompt diversity is the primary bottleneck for Long-CoT distillation.

\subsection{The Impact of CoT Correctness}
\label{sec:pvsn}

\paragraph{Setup.}
To systematically isolate the impact of factual correctness, we establish a controlled training environment. We randomly sample 3,000 prompts from the DAPO-Math-17K~\citep{yu2025dapo} dataset.
For each prompt, we employ Qwen3-4B (non-thinking) to generate 16 reasoning trajectories, yielding a pool of 48,000 question-answer pairs.
Subsequently, we utilize math-verify~\citep{huggingface2024mathverify} to rigorously assess the correctness of each trajectory.
This evaluation yields 22,222 correct (positive) samples and 25,778 incorrect (negative) samples.
This natural distribution provides a well-balanced foundation for comparison.
Finally, we independently fine-tune the base models on these two disjoint datasets.
\begin{figure}[htbp]
    \centering
    \includegraphics[width=\textwidth]{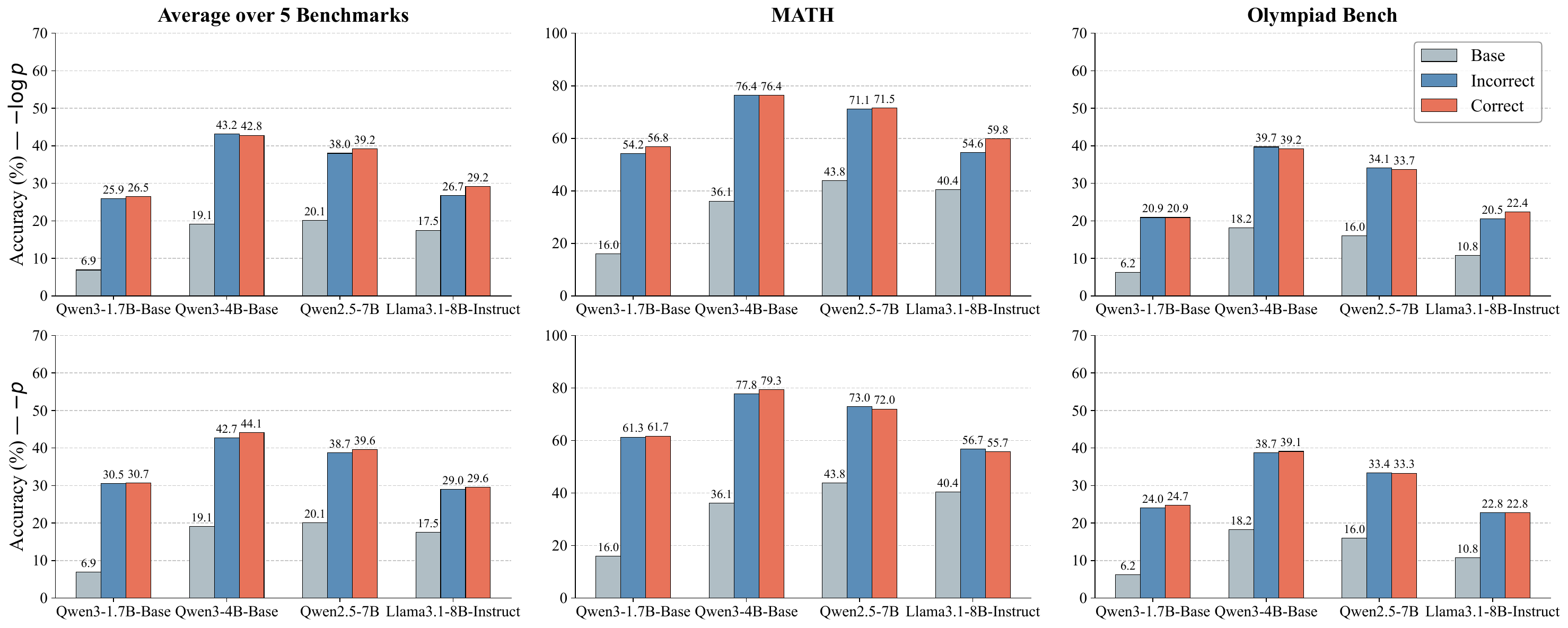}
    \caption{\textbf{Incorrect Trajectories Teach Long-CoT Behavior Almost as Effectively as Correct Ones.}
We compare models fine-tuned on positive versus negative CoT trajectories under two training objectives.
Across models and benchmarks, negative trajectories retain most of the distillation gains achieved by positive trajectories.
}
    \label{fig:Pos_vs_Neg}
\end{figure}

\begin{table}[ht]
\centering
\small
\caption{\textbf{The Negligible Impact of CoT Correctness Extends Beyond Math Benchmarks.} 
We evaluate whether the weak dependence on CoT correctness generalizes to non-math domains, including general reasoning, scientific reasoning, and coding tasks.
Across models and training objectives, negative CoT trajectories achieve average performance nearly identical to positive trajectories. P and N denote positive and negative CoT trajectories, respectively.}
\label{tab:pvsn}
\resizebox{0.95\textwidth}{!}{
\setlength{\tabcolsep}{4pt} 
\begin{tabular}{lccccl|lccccl}
\toprule
\multirow{2}{*}{Method} & \multirow{2}{*}{Data} & \multicolumn{4}{c|}{\textbf{Qwen3-1.7B-Base}} & \multirow{2}{*}{Method} & \multirow{2}{*}{Data} & \multicolumn{4}{c}{\textbf{Qwen3-4B-Base}} \\ \cmidrule(lr){3-6} \cmidrule(l){9-12}
 & & MMLUPro & GPQA-D & LCBv2 & Average & & & MMLUPro & GPQA-D & LCBv2 & Average \\ \midrule
Base & - & 28.1 & 14.4 & 8.9 & 17.1 & Base & - & 46.1 & 16.2 & 19.8 & 27.4 \\
\rowcolor{gcolor1}
-logp & P & 35.3 & 21.3 & 13.8 & 23.5 & -logp & P & 52.9 & 29.6 & 29.4 & 37.3 \\
\rowcolor{gcolor1}
-logp & N & 34.3 & 23.0 & 14.3 & 23.9 \posdiff{0.4} & -logp & N & 56.4 & 31.5 & 28.8 & 38.9 \posdiff{1.6} \\
\rowcolor{gcolor2}
-p & P & 34.9 & 15.7 & 9.3 & 20.0 & -p & P & 54.3 & 34.3 & 24.8 & 37.8 \\
\rowcolor{gcolor2}
-p & N & 34.7 & 14.7 & 7.9 & 19.1 \negdiff{0.9} & -p & N & 52.6 & 33.5 & 24.4 & 36.8 \negdiff{1.0} \\ \midrule
\multirow{2}{*}{Method} & \multirow{2}{*}{Data} & \multicolumn{4}{c|}{\textbf{Llama-3.1-8B-Instruct}} & \multirow{2}{*}{Method} & \multirow{2}{*}{Data} & \multicolumn{4}{c}{\textbf{Qwen2.5-7B}} \\ \cmidrule(lr){3-6} \cmidrule(l){9-12}
 & & MMLUPro & GPQA-D & LCBv2 & Average & & & MMLUPro & GPQA-D & LCBv2 & Average \\ \midrule
Base & - & 34.9 & 24.5 & 17.7 & 25.7 & Base & - & 38.2 & 27.2 & 5.6 & 23.7 \\
\rowcolor{gcolor1}
-logp & P & 42.3 & 32.7 & 21.5 & 32.2 & -logp & P & 50.6 & 24.0 & 29.4 & 34.7 \\
\rowcolor{gcolor1}
-logp & N & 40.4 & 32.1 & 21.3 & 31.3 \negdiff{0.9} & -logp & N & 48.7 & 26.7 & 28.0 & 34.5 \negdiff{0.2} \\
\rowcolor{gcolor2}
-p & P & 42.3 & 22.3 & 20.9 & 28.5 & -p & P & 47.7 & 22.1 & 31.2 & 33.7 \\
\rowcolor{gcolor2}
-p & N & 41.9 & 23.4 & 20.7 & 28.7 \posdiff{0.2} & -p & N & 45.9 & 22.0 & 30.1 & 32.7 \negdiff{1.0} \\ \bottomrule
\end{tabular}
}
\end{table}

\paragraph{Results.}
As illustrated in \Cref{fig:Pos_vs_Neg}, models fine-tuned exclusively on negative (incorrect) reasoning trajectories exhibit remarkable performance gains on mathematical reasoning tasks, capturing the vast majority of the improvements achieved by their positive (correct) counterparts. Quantitatively, on mathematical datasets, models trained on negative samples achieve \textbf{97.7\%} of the absolute accuracy of those trained on positive samples (${Acc}_{\text{neg}} / Acc_{\text{pos}}$) and retain an impressive \textbf{94.7\%} of the relative performance gain ($({Acc}_{\text{neg}} - {Acc}_{\text{base}}) / ({Acc}_{\text{pos}} - {Acc}_{\text{base}})$).
This trend holds consistently across different base models and under both $-\log p$ and $-p$ loss, resulting in a consistently narrow performance gap between the positive and negative data settings.
Furthermore, \Cref{tab:pvsn} demonstrates that this weak dependence on CoT correctness extends well beyond the mathematical domain. Across broader evaluation suites encompassing general reasoning (MMLU-Pro), scientific problem-solving (GPQA-Diamond), and coding (LiveCodeBench-v2), the average performance discrepancy between positive and negative CoT distillation remains strikingly minimal, averaging within approximately a 1\% margin.
Additional evaluations on AIME 2026 and LiveCodeBench-v6, along with models trained on science QA and with alternative teachers, show the same positive-versus-negative pattern (\Cref{app:rebuttal}). Collectively, these results suggest that final-answer correctness is not the dominant factor in the settings studied.

\subsection{The Utility of Incorrect Trajectories}
\label{sec:pwithn}

\paragraph{Setup.}
To investigate how the ratio of positive to negative samples influences distillation outcomes, we leverage the datasets constructed in \Cref{sec:pvsn}. Specifically, we establish varying baselines of positive samples by subsampling subsets of $\{512, 1024, 2048, 5120\}$ correct trajectories.
For each baseline size, we then supplement the training data with incorrect trajectories at positive-to-negative ratios of $\{\text{1:0, 1:1, 1:2, 1:5}\}$.
To ensure that the models can fully learn and converge, especially in scenarios where the total dataset size is exceedingly small (e.g., 512 samples), we extend the training duration to 15 epochs across all mixed datasets.

\begin{figure}[htbp]
    \centering
    \includegraphics[width=0.95\linewidth]{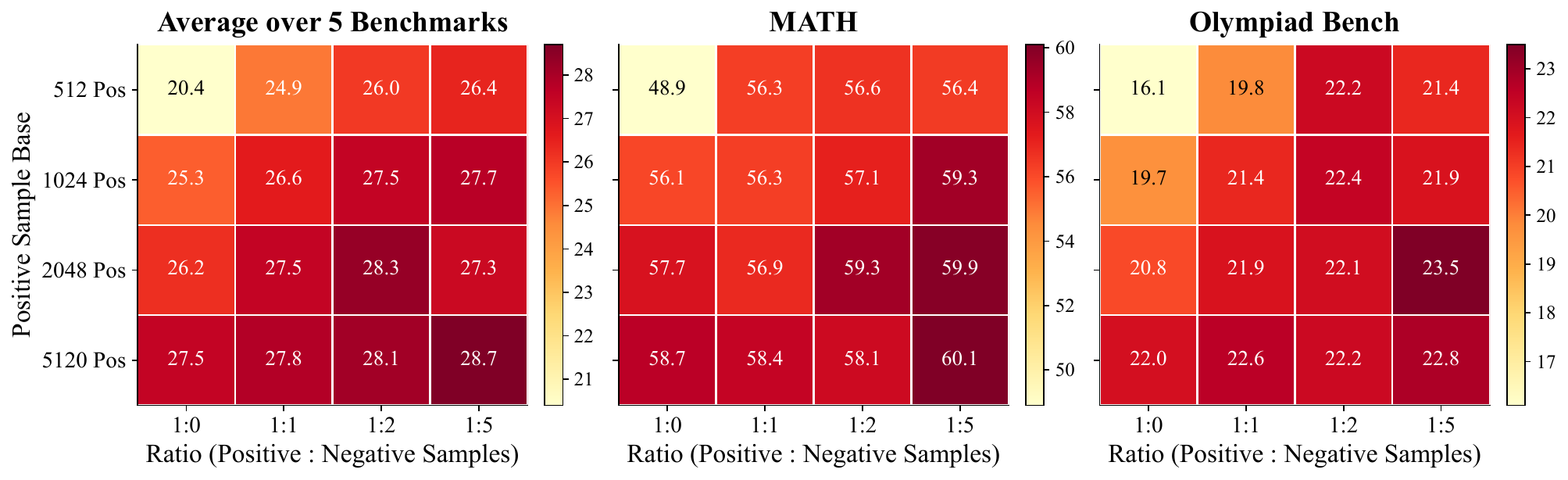}
    \vspace{-5pt}
    \caption{
    \textbf{Negative CoTs Effectively Mitigate the Scarcity of Correct Data.}
    We train Qwen3-1.7B-Base with different positive sample bases and varying ratios of positive to negative CoT trajectories.
    Adding negative trajectories consistently improves performance when positive data is scarce.
    }
    \label{fig:mixed-positive-negative}
    \vspace{-5pt}
\end{figure}

\paragraph{Results.}
As illustrated in \Cref{fig:mixed-positive-negative}, injecting negative trajectories consistently enhances performance when positive data is scarce.
For instance, when the positive sample base is strictly limited to 512, increasing the proportion of negative samples from 1:0 to 1:5 yields a substantial absolute increase in the average accuracy across five math benchmarks, rising from 20.4\% to 26.4\%.
However, the benefit of incorporating incorrect trajectories diminishes as the volume of positive data grows.
Ultimately, these findings confirm that in scenarios where correct samples are scarce, supplementing the training set with mixed incorrect trajectories effectively bolsters the model's acquisition of reasoning patterns.
This reinforces the conclusion that incorrect CoT trajectories still provide vital behavioral supervision for Long-CoT distillation.

\subsection{The Impact of Prompt Diversity}
\paragraph{Setup.}
To systematically isolate and evaluate the impact of prompt diversity on Long-CoT distillation, we design a controlled experiment where the total trajectory budget remains fixed.
Specifically, we randomly sample prompt subsets of varying sizes from the DAPO-Math-17K dataset, containing $\{\text{500, 1K, 2K, 4K, 8K, 16K}\}$ unique prompts. To ensure the total volume of training samples is strictly controlled within the same order of magnitude across all settings, we employ Qwen3-4B (non-thinking) to sample an inversely proportional number of candidate trajectories for each prompt: generating $\{\text{128, 64, 32, 16, 8, 4}\}$ responses per prompt, respectively. Finally, we apply standard rejection sampling to these generated candidates, filtering strictly for correct reasoning traces to construct our final fine-tuning datasets.

\begin{figure}[htbp]
    \centering
    \includegraphics[width=\textwidth]{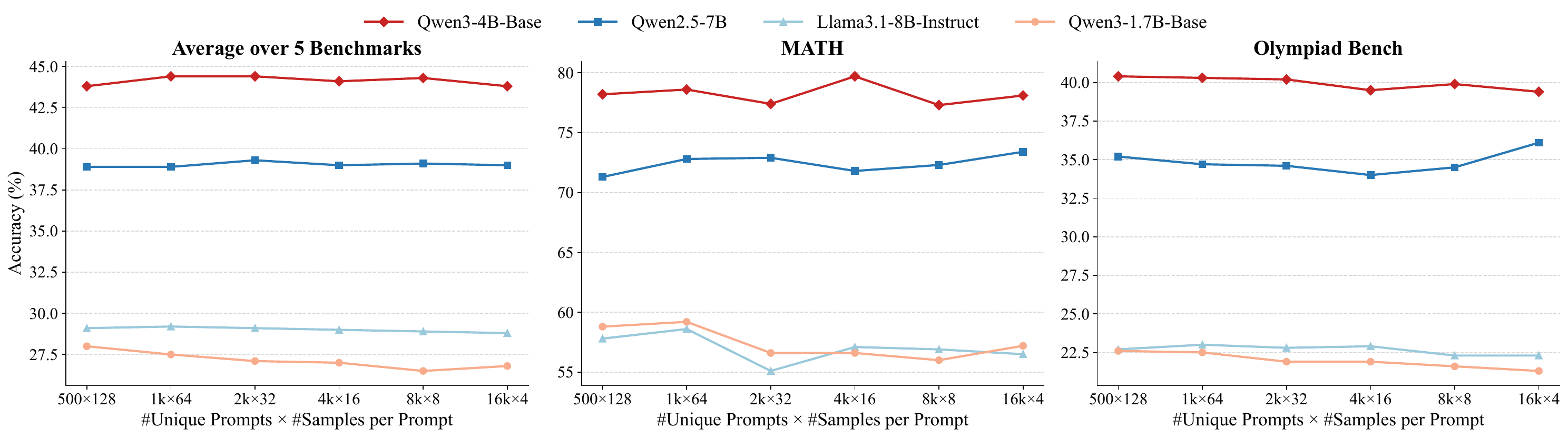}
    \caption{
\textbf{Prompt Diversity Is Not the Bottleneck Under a Fixed Trajectory Budget.}
We fix the total number of CoT trajectories and vary the source prompt diversity, ranging from a small prompt set with many sampled trajectories per prompt ($500 \times 128$) to a large prompt set with fewer trajectories per prompt ($16\mathrm{k} \times 4$).
    }
    \label{fig:prompt_diversity}
\end{figure}

\paragraph{Results.}
As depicted in \Cref{fig:prompt_diversity}, maintaining a fixed trajectory budget reveals that the reasoning performance of the distilled models is remarkably insensitive to the diversity of the source prompts. Across four distinct base models and five mathematical benchmarks, the average accuracy curves remain notably flat as the data composition transitions from a highly diverse prompt set to a heavily restricted one. For instance, fine-tuning on merely 500 unique problems sampled 128 times each, yields average benchmark results virtually identical to those achieved by utilizing 16K distinct problems sampled 4 times each.
This stable performance trajectory suggests that broad problem coverage is not the dominant bottleneck at the trajectory budgets considered.

\section{The Behavioral Hypothesis of Long-CoT Distillation}
\label{sec:behavioral_hypothesis}

Our observations in \Cref{sec:findings} suggest that neither final-answer correctness nor broad prompt diversity is the dominant bottleneck in the studied settings. We propose the \textit{Behavioral Hypothesis}: a central benefit of Long-CoT distillation is learning structured reasoning patterns rather than memorizing perfectly correct, diverse problem-specific facts. 

We distinguish two complementary aspects of a reasoning trajectory, \textit{Factual Knowledge} and \textit{Reasoning Behaviors} (illustrated in \Cref{fig:case_study}).

\subsection{Deconstructing CoT: Factual Knowledge vs. Reasoning Behaviors}

\textbf{Factual Knowledge.} This dimension encompasses domain-specific axioms, problem constraints, and the precise execution of intermediate computations. Conventional distillation pipelines implicitly assume that models must internalize a broad, error-free spectrum of this knowledge. However, while a localized factual flaw (e.g., a hallucinated index or arithmetic error) almost inevitably corrupts the final answer, it does not necessarily invalidate the structural integrity of the entire trajectory.

\textbf{Reasoning Behaviors.} Distinct from factual accuracy, this dimension governs the meta-cognitive processes of problem-solving. It dictates how a model organizes its \textit{pre-existing} knowledge into a deep, step-by-step format. It includes domain-agnostic, highly reusable actions such as problem decomposition, logical chaining, and sustaining derivation over long contexts. As demonstrated by our negative sample training (\Cref{fig:case_study}), a model can execute flawless reasoning behaviors even when operating on flawed initial facts.

Building on this distinction, we test whether \textbf{behavioral alignment} can explain gains beyond problem-specific knowledge injection using two complementary comparisons.

\subsection{Evidence I: Resistance to Factual Memorization}


\begin{wraptable}{r}{0.4\textwidth}
    \vspace{-3.5em}
    \centering
    \caption{\textbf{Models Avoid Overfitting to Flawed Facts on the Training Set.} Numbers in parentheses indicate performance gains relative to the base model.}
    \label{tab:training_memorization}
    \setlength{\tabcolsep}{3pt}
    \resizebox{\linewidth}{!}{
    \begin{tabular}{llll}
        \toprule
        \multirow{2}{*}{\textbf{Model}} & \multirow{2}{*}{\textbf{Loss}} & \multicolumn{2}{c}{\textbf{Acc (Gains) (\%)}} \\
        \cmidrule(lr){3-4}
        & & \textbf{Positive} & \textbf{Negative} \\
        \midrule
        \multirow{2}{*}{\begin{tabular}[c]{@{}l@{}}Qwen3\\1.7B-Base\end{tabular}}  
        & -logp & 14.3 \posdiff{7.4} & 13.5 \posdiff{6.6} \\
        & -p & 17.3 \posdiff{10.4} & 15.7 \posdiff{8.8} \\
        \midrule
        \multirow{2}{*}{\begin{tabular}[c]{@{}l@{}}Qwen3\\4B-Base\end{tabular}}  
        & -logp & 36.4 \posdiff{23.3} & 34.6 \posdiff{21.5} \\
        & -p & 35.4 \posdiff{22.3} & 34.1 \posdiff{21.0} \\
        \bottomrule
    \end{tabular}
    }
    \vspace{-25pt}
\end{wraptable}

\paragraph{Setup.}
To assess the degree to which models memorize flawed factual details, we evaluate those fine-tuned on incorrect trajectories directly on their respective training sets in \Cref{tab:training_memorization}. The training configuration and datasets are identical to \Cref{sec:pvsn}.

\paragraph{Results.}
We observe a striking phenomenon: fine-tuning on negative samples yields significant improvements even when evaluated on the training set itself. If the distillation process relied heavily on factual knowledge acquisition, a model trained on flawed facts would predictably rote-memorize those errors, causing its ground-truth accuracy on the training set to collapse. 
Instead, both Qwen3-1.7B-Base and Qwen3-4B-Base maintain highly comparable accuracy to their positive counterparts across both SFT objectives. This small divergence is consistent with models learning useful structural patterns from flawed trajectories rather than simply reproducing the erroneous final answers.

\subsection{Evidence II: Isolating Behavioral Alignment via Solvable Problems}

\paragraph{Setup.}
To explicitly isolate the behavioral mechanism from knowledge transfer, we mitigate the impact of ``novel knowledge injection''. We construct training datasets exclusively comprising problems that the base models can already solve. 
Specifically, we evaluate the models on the DAPO-Math-17K dataset across 8 sampling iterations, retaining prompts correctly answered at least 7 times. This yields highly confident, solvable subsets (Easy-444 for Qwen3-8B-Base and Easy-643 for Qwen3-14B-Base).
Since the base model already solves these prompts reliably, this comparison limits the role of novel problem-specific knowledge as an explanation for subsequent gains.
To ensure a fair volumetric comparison with the experiments in \Cref{sec:pvsn}, we repeatedly sample trajectories from these subsets to match the scale of our baseline. We then benchmark these models against those fine-tuned on the Random-3k dataset.

\begin{figure}[htbp]
    \centering
    \includegraphics[width=0.95\linewidth]{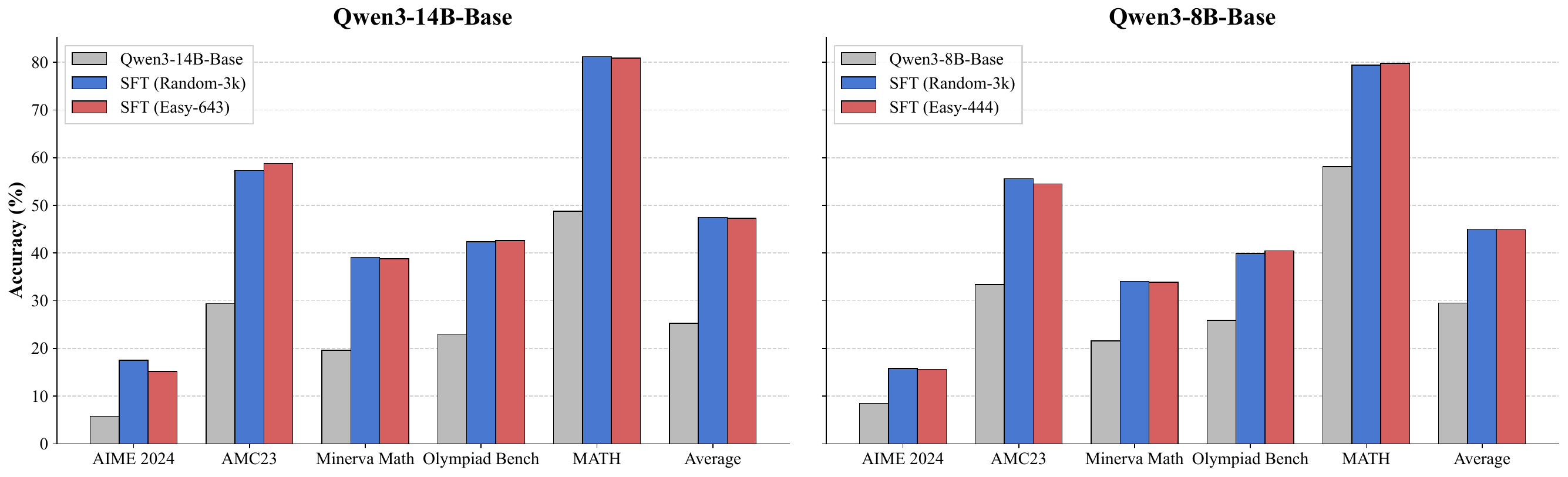}
    \caption{\textbf{Substantial Gains on Previously Solvable Problems.} Fine-tuning on easy subsets that the base model can already solve (Easy-643/444) yields improvements comparable to training on a larger, randomly sampled dataset (Random-3k) across math benchmarks. This reduces the role of new problem-specific answers as an explanation for the gains.}
    \label{fig:easy_subset}
\end{figure}

\paragraph{Results.}
As illustrated in \Cref{fig:easy_subset}, fine-tuning base models exclusively on subsets they can already solve yields remarkable improvements that closely match the Random-3k baseline across all evaluated benchmarks. Because the base models could already answer these prompts, learning previously unseen problem-specific answers alone is unlikely to explain the gains. 

Remarkably, despite being exposed to extremely limited novel factual knowledge, the models still perform competitively with the performance of the broader SFT baseline.
Together with \Cref{tab:training_memorization} and the late-layer representation comparison in \Cref{app:rebuttal}, these results support the \textit{Behavioral Hypothesis}: distillation helps models organize what they already know into reusable reasoning patterns.

\section{Further Exploration: Harmonizing CoT Complexity and Optimization}
\label{sec:loss}
Motivated by the role of structured reasoning, we investigate how trajectory complexity interacts with the choice of optimization objective. Specifically, we compare the standard cross-entropy loss ($-\log p$) with the recently popularized plain probability-based loss ($-p$) across datasets of varying reasoning depths.

\subsection{Empirical Divergence Across Complexity Levels}
\paragraph{Setup.}
To systematically study the effect of CoT complexity, we construct three datasets representing distinct tiers of reasoning depth: Simple, Medium, and Complex.
We repurpose the data from \Cref{sec:pvsn} for the Medium dataset.
For the Complex dataset, we employ the same pipeline but run Qwen3-4B (thinking), sampling 8 responses per prompt. The Simple dataset is built by sampling a subset from NuminaMath-CoT~\citep{li2024numinamath}.

The token length distributions and word cloud analyses (detailed in \Cref{app:dataset_stats}) highlight the significant semantic contrast across these tiers.
The average token length scales aggressively across the Simple (426), Medium (1,337), and Complex (9,062) datasets.
Crucially, the word cloud analysis reveals that this scaling in length is accompanied by a qualitative shift in reasoning behaviors.
While the Simple dataset is dominated by concrete, task-oriented vocabulary (e.g., "solution", "value"), the Complex dataset is heavily populated with meta-cognitive markers (e.g., "wait", "maybe", "alternatively"). This demonstrates that the extended trajectories in the Complex dataset are not merely bloated calculations, but encapsulate genuine trial-and-error and advanced self-correction behaviors.


\begin{figure}[htbp]
    \centering
    \includegraphics[width=\textwidth]{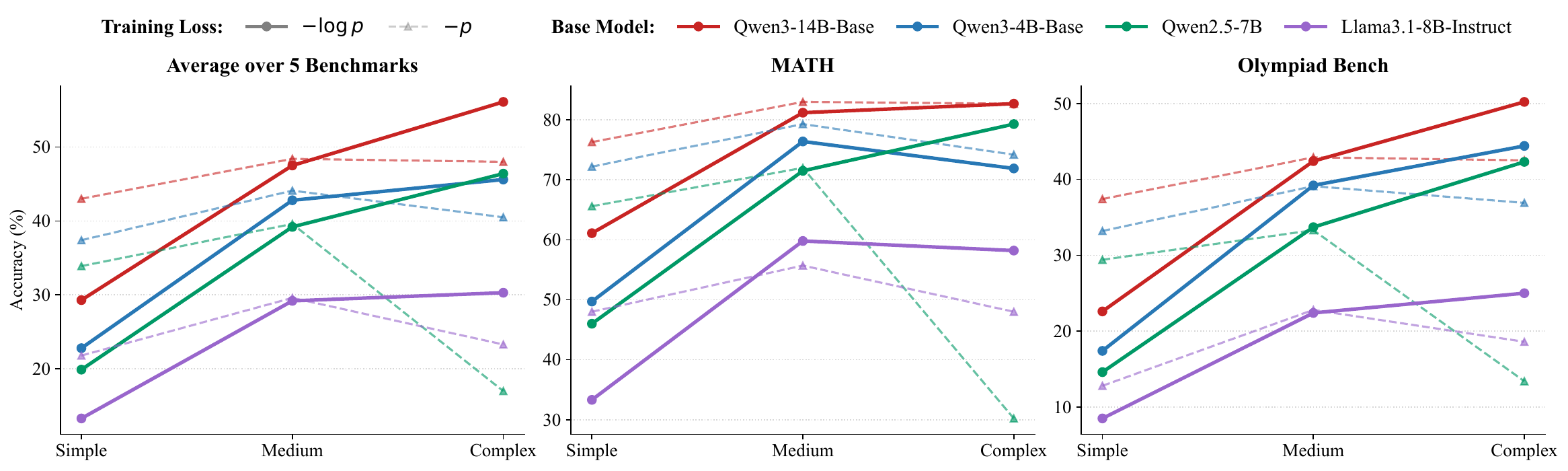}
    \caption{
        \textbf{CoT Complexity Dictates the Appropriate Distillation Objective.}
        We compare four base models fine-tuned with $-\log p$ loss (solid lines) and $-p$ loss (dashed lines) across varying CoT complexities. The results reveal a clear dichotomy: while the $-p$ loss excels on simpler CoTs, it suffers from pronounced degradation on highly complex reasoning traces. Conversely, the standard $-\log p$ loss scales robustly with complexity.
    }
    \label{fig:loss_complexity}
\end{figure}

\paragraph{Results.}
As illustrated in \Cref{fig:loss_complexity}, our evaluation reveals a clear divergence driven by CoT complexity. First, we find that training on more complex reasoning data unlocks higher peak performance.
However, this scaling benefit is conditional on using the correct loss function.
Specifically, this divergence directly challenges the emerging consensus that the $-p$ loss is universally superior for reasoning acquisition~\citep{wu2026on,li2025beyond}.
As depicted in \Cref{fig:loss_complexity}, while the $-p$ loss indeed yields notable gains on simpler CoTs, this initial advantage completely collapses as reasoning traces become more intricate.
On the Complex dataset, models trained with $-p$ suffer a steep performance drop.
In contrast, the standard $-\log p$ loss scales robustly, maintaining steady upward trajectories and effectively capturing the benefits of extended reasoning.
Ultimately, these findings highlight that the efficacy of these distillation objectives is highly context-dependent, underscoring the necessity of matching the loss function to the intrinsic complexity of the training trajectories.

\subsection{The Mechanics: A Perspective on Gradient Dynamics}

To understand this performance divergence, we revisit the gradient dynamics of the two objectives. Taking the derivative of the loss with respect to the predicted probability $p$ yields $\frac{\partial}{\partial p}(-\log p)=-\frac{1}{p}$ and $\frac{\partial}{\partial p}(-p)=-1$.
Applying the chain rule, the gradient of $-\log p$ loss with respect to the model parameters $\theta$ can be formally expressed directly in terms of $-p$:
\begin{equation}
\label{equ:grad}
    \nabla_\theta (-\log p) = \frac{1}{p} \nabla_\theta (-p).
\end{equation}
This highlights a core difference in how the objectives treat "hard tokens" (where prediction confidence $p$ is small).
The $-\log p$ loss acts as a dynamic amplifier, heavily penalizing low-$p$ predictions via the $1/p$ multiplier.
In contrast, the $-p$ loss applies a uniform penalty, resulting in the \textit{selective under-learning} of hard tokens.
This mechanism drives the performance divergence across different complexity tiers.

This selective learning mechanism explains the performance crossing across complexity tiers.
\textit{Simple CoTs: The Overfitting Dilemma.} In straightforward reasoning, the overall logic is highly predictable.
Hard tokens are relatively rare and typically represent random noise rather than key logical transitions.
Here, the $1/p$ amplifier of $-\log p$ forces the model to over-optimize on this noise, leading to overfitting.
Conversely, the $-p$ loss avoids excessive updates on these low-$p$ regions, acting as a robust regularizer that improves generalization.
\textit{Complex CoTs: The Coherence Collapse.} Complex reasoning involves behaviors like backtracking and hypothesis exploration, which drastically increase the density of hard tokens.
Because $-p$ under-penalizes low-$p$ predictions, it fails to provide the rigorous supervision necessary to master these vital transitions.
Since deep reasoning is strictly sequential, failing to internalize these structural tokens corrupts the logical coherence of the entire trajectory, fracturing the reasoning chain.
The response lengths in \Cref{tab:loss_length} are consistent with this account: after training on complex trajectories averaging roughly 9K tokens, $-p$-trained Qwen2.5-7B and Llama-3.1-8B-Instruct generate more than 20K tokens on average, versus roughly 14K and 13K under $-\log p$, respectively. This contrast suggests that learning difficult transitions, including when to stop, matters for long-chain coherence.
In summary, while the $-p$ objective prevents overfitting on simple samples, the gradient amplification intrinsic to $-\log p$ is essential for maintaining structural coherence in complex, long-chain reasoning.

\section{Related Work}
Long-CoT distillation has become an important method for endowing models with the ability to reason about complex tasks~\citep{guan2025rstar}.
Early approaches focused on constructing training sets by collecting large and diverse amounts of data through a pipeline.
\citet{guo2025deepseek} constructed a dataset containing 600K CoT instances through rejection sampling to train other models.
Similarly, \citet{zhang2025thyme} constructed a 500k dataset encompassing various scenarios.
Furthermore, \citet{luo2025deconstructing} emphasized the correctness of CoT, optimizing the CoT dataset by simplifying and correcting the intermediate steps.
Alternatively, \citet{zhao2026mind} propose to rewrite the SFT dataset, arguing that data diversity is one of the key factors.
These pipelines motivate controlled tests of how correctness and prompt coverage affect distillation. Earlier work already found that flawed traces can help: \citet{li2025llms} contrast CoT structure with perturbed content, and \citet{li2024turning} use negative traces with dual-LoRA and calibration. \citet{du2025challenge} demonstrate reasoning gains from just 20--50 high-quality CoTs. Unlike those settings, we test direct SFT on flawed traces across model families, fixed-budget prompt diversity, and the interaction of CoT complexity with the training loss.

Another line of work~\citep{zhu2026proximal,liu2026profit,chen2026sed} focuses on modifying the optimization objective of SFT to enhance post-training reasoning capabilities.
\citet{wu2026on} unified SFT and reinforcement learning (RL) under a single framework and proposed that using $-p$ instead of $-\log p$ as the loss function can improve generalization on reasoning tasks.
Subsequently, \citet{zhu2026anchored,li2025beyond} adopted this approach, incorporating $-p$ as part of their loss function.
However, our work reveals that the effectiveness of this modified loss function in prior studies is limited to relatively simple SFT datasets. On longer and more complex datasets, the standard SFT loss performs better in our experiments. Complementarily, \citet{fan2026saft} interpolate between the two losses according to data noise; our study identifies trajectory complexity as another relevant factor.

Even though CoT distillation has proven to be widely effective, the specific components that actually drive its success are not yet fully understood~\citep{yu2025reassessing}.
\citet{li2025llms} argued that the structure of CoT is crucial for model learning.
On the other hand, \citet{chandra2025shape} contended that the underlying data distribution is of paramount importance.
Furthermore, \citet{feng2025characterizes} argued that both naive CoT lengthening and increased review steps actually hurt model performance.
\citet{ren2026rethinking} re-examined the generalization of SFT from the perspectives of the optimization process, data, and model capabilities, concluding that SFT can indeed achieve generalization.
Building on these observations, we compare multiple architectures, evaluation domains, losses, and training-data compositions; we also test whether flawed traces can supplement scarce correct data.
\section{Conclusion}
This study dismantles the established assumptions of Long-CoT distillation, proving that successful reasoning transfer does not strictly require flawless trajectories or extensive prompt diversity. Instead, we introduce the Behavioral Hypothesis: distillation fundamentally teaches models the structural methodology of step-by-step reasoning, rather than rote factual memorization. Furthermore, we reveal a critical divergence in optimization: while the $-p$ loss excels on simpler trajectories, the standard cross-entropy loss ($-\log p$) remains strictly essential for scaling to complex, lengthy reasoning. Ultimately, by isolating these behavioral drivers, our findings provide a highly efficient and principled blueprint for data curation and objective selection in the next generation of reasoning models.



\bibliographystyle{unsrtnat}
\bibliography{main}

\appendix
\newpage
\section{Experimental Details}

\subsection{Losses}
\paragraph{Unifying the two objectives.}
To systematically analyze the two SFT losses ($-\log p$ and $-p$), it is essential to recognize that they are not isolated heuristics. Instead, they can be elegantly unified under a generalized family of probability-based loss functions parameterized by \(\alpha\), defined as \(f_\alpha(p) = \frac{1 - p^\alpha}{\alpha}\). This formulation bridges the two objectives:
\begin{equation}
\label{equ:form}
    f_\alpha(p) = 
    \begin{cases} 
    \lim_{\alpha \to 0} \frac{1 - p^\alpha}{\alpha} = -\log p, & \text{(NLL loss)}, \\[8pt]
    \frac{1 - p^1}{1} = 1 - p \equiv -p, & \text{(plain-}p \text{ loss, up to a constant)}.
    \end{cases}
\end{equation}

While \Cref{equ:form} demonstrates that \(-\log p\) and \(-p\) share a unified mathematical origin, our experiments reveal that their actual optimization behaviors diverge profoundly depending on the complexity of the reasoning task. We will extensively evaluate their empirical performance and fundamentally deconstruct their gradient dynamics in \Cref{sec:loss}.

\begin{figure}[htbp]
    \centering
    \small
    
    \fcolorbox{black!20}{yellow!5}{
        \begin{minipage}{0.96\textwidth}
            \vspace{1mm} 
            \textbf{Question:}\\
            Find the value of the product 
            \[ \frac{8}{4} \cdot \frac{12}{8} \cdot \frac{16}{12} \cdots \frac{4n + 4}{4n} \cdots \frac{2008}{2004}. \]
            \vspace{-2mm} 
        \end{minipage}
    }
    
    \vspace{2mm} 
    
    \begin{tabularx}{\textwidth}{@{}X|X@{}}
        \toprule
        \textbf{\textcolor{blue!80!black}{\ding{51} Positive Sample}} & 
        \textbf{\textcolor{red!80!black}{\ding{55} Negative Sample}} \\
        \textit{(Correct Fact \& Perfect Behavior)} & 
        \textit{(Flawed Fact, but Same Behavior)} \\
        \midrule
        
        \textbf{Step 1: Understand the pattern} \newline
        The terms are: $\frac{8}{4}, \frac{12}{8}, \dots, \frac{2008}{2004}$. \newline
        Each term has the general form:
        \[ \frac{4(n+1)}{4n} = \frac{n+1}{n} \]
        So the product becomes:
        \[ \prod_{n=1}^{501} \frac{n+1}{n} \]
        \textcolor{gray}{\textit{--- Correct boundary condition ---}} & 
        
        \textbf{Step 1: Understand the pattern} \newline
        The terms are: $\frac{8}{4}, \frac{12}{8}, \dots, \frac{2008}{2004}$. \newline
        Let the general term be:
        \[ \frac{4(k+1)}{4k} = \frac{k+1}{k} \]
        So the product becomes:
        \[ \prod_{k=2}^{502} \frac{k+1}{k} \]
        \textcolor{red}{\textbf{[!] Factual Error:}} Hallucinated starting index ($k=2$ instead of $k=1$). \\
        
        \midrule
        
        \textbf{Step 2: Telescoping Product} \newline
        Write out the product explicitly:
        \[ \frac{2}{1} \cdot \frac{3}{2} \cdot \frac{4}{3} \cdots \frac{502}{501} \]
        Observe that this is a \textbf{telescoping product}. Most terms cancel out... & 
        
        \textbf{Step 2: Telescoping Product} \newline
        This is a \textbf{telescoping product}. \newline
        Let's write out the product:
        \[ \frac{3}{2} \cdot \frac{4}{3} \cdot \frac{5}{4} \cdots \frac{503}{502} \]
        We can see that most terms cancel out... \\
        
        \midrule
        
        \textbf{Final Answer:} \newline
        \[ \boxed{502} \] & 
        
        \textbf{Final Answer:} \newline
        \[ \boxed{\frac{503}{2}} \] \\
        
        \bottomrule
    \end{tabularx}
    
    \vspace{3mm} 
    
    \fcolorbox{black!20}{gray!5}{
        \begin{minipage}{0.96\textwidth}
            \vspace{1mm} 
            \textbf{Shared Reasoning Behaviors (The Essence of Distillation):}
            \begin{itemize}
                \item \textbf{Pattern Recognition:} Successfully identifies the complex "telescoping product" structure.
                \item \textbf{Process Unfolding:} Expands the abstract symbols into explicit sequential multiplications without skipping steps.
                \item \textbf{Logical Deduction:} Accurately describes the continuous "cancel out" mechanism between numerators and denominators.
            \end{itemize}
            \vspace{0mm} 
        \end{minipage}
    }
    
    \caption{\textbf{Factual Flaws vs. Behavioral Alignment in Long-CoT Distillation.} Given the same initial problem, the negative sample hallucinated an incorrect starting index ($k=2$), leading to a wrong final answer, yet it perfectly replicated the core, step-by-step reasoning behaviors of the positive sample. This structurally intact deduction demonstrates why incorrect CoT trajectories still provide high-quality behavioral supervision.}
    \label{fig:case_study}
\end{figure}

\subsection{Dataset Demonstration and Analysis}
\label{app:dataset_stats}
\paragraph{A Concrete Case Study.}
To concretely illustrate the distinction between factual knowledge and reasoning behaviors, \Cref{fig:case_study} presents a side-by-side comparison of a positive and a negative reasoning trajectory for the same mathematical problem. In the positive sample, the model correctly identifies the boundary conditions and executes flawless calculations to reach the accurate final answer. Conversely, the negative sample introduces a localized factual error early in the process---specifically, hallucinating an incorrect starting index ($k=2$ instead of $k=1$). Inevitably, this flawed initial fact corrupts the final calculated answer. 

However, crucially, the structural integrity of the reasoning process remains completely intact. Despite the factual hallucination, the negative sample successfully exhibits advanced reasoning behaviors identical to those of the positive sample: it correctly recognizes the complex "telescoping product" pattern, unfolds the abstract symbols into explicit sequential multiplications without skipping steps, and logically deduces the continuous "cancel out" mechanism. This concrete example vividly demonstrates why incorrect trajectories can still provide high-quality behavioral supervision for Long-CoT distillation: they effectively teach the model the structured methodology of \textit{how} to reason step-by-step, even when operating on flawed initial facts.

\begin{figure}[htbp]
    \centering
    \includegraphics[width=0.5\textwidth]{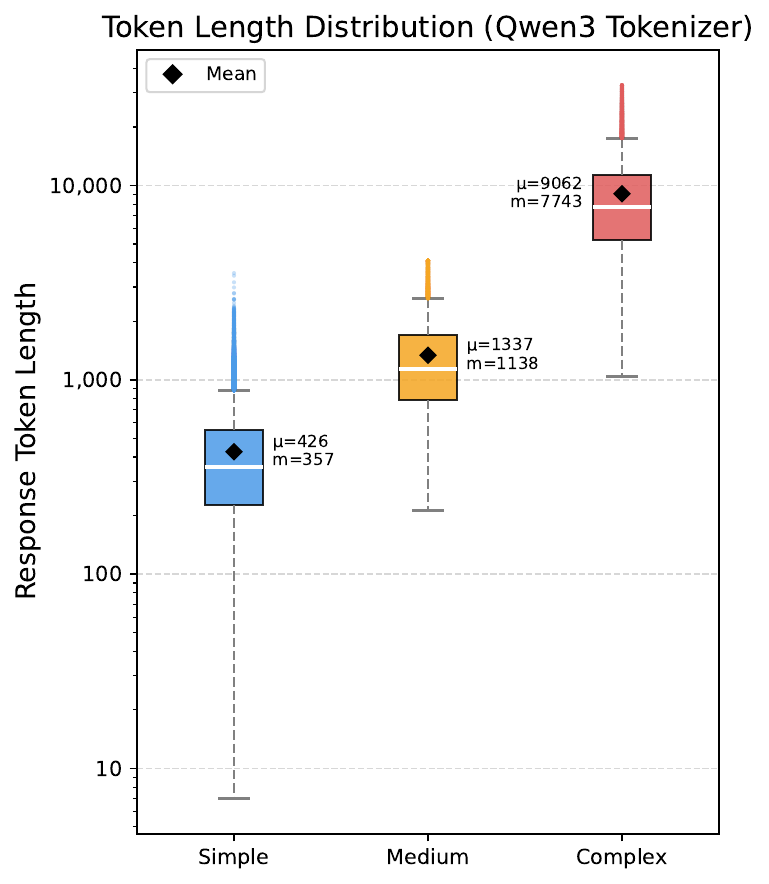}
    \caption{Token Length Distribution Across Reasoning Complexity Tiers. The boxplot illustrates the scaling of response lengths across Simple, Medium, and Complex datasets.}
    \label{fig:token_boxplot}
\end{figure}

\paragraph{Datasets with Different Complexity Tiers.}
To provide a deeper understanding of the intrinsic differences across the varying complexity levels, this section presents detailed dataset statistics and analysis. First, as illustrated by the boxplot in \Cref{fig:token_boxplot}, the response token lengths scale aggressively across the Simple, Medium, and Complex datasets. This provides a quantitative reflection of the progressively increasing reasoning depth. 

\begin{figure}[htbp]
    \centering
    \includegraphics[width=\textwidth]{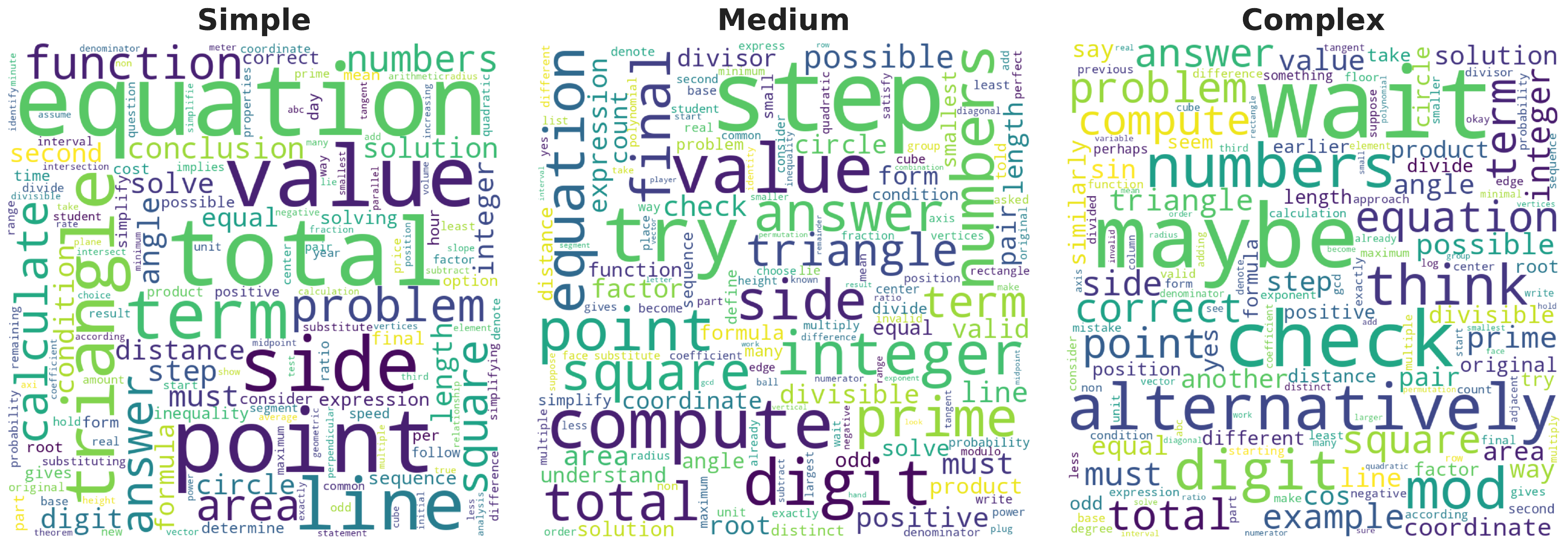}
    \caption{Word Cloud Analysis Reveals Lexical Shifts Across CoT Complexities. The lexical distribution shows a transition from concrete task-oriented words in the Simple dataset, through exploratory terms in the Medium dataset, to meta-cognitive markers in the Complex dataset.}
    \label{fig:wordcloud}
\end{figure}

However, this scaling in complexity is not merely a superficial accumulation of token length. As further revealed by the word cloud analysis in \Cref{fig:wordcloud}, there is a fundamental qualitative shift in the lexical distribution across the datasets. In the Simple dataset, the vocabulary is heavily dominated by concrete, task-oriented words (e.g., "function", "equation", "value"). Interestingly, the Medium dataset serves as a clear transitional phase: alongside standard mathematical terms, we begin to observe the early emergence of exploratory vocabulary such as "try" and "possible". Ultimately, in stark contrast to the simplest instances, the Complex dataset exhibits a dense population of advanced meta-cognitive markers (e.g., "wait", "maybe", "alternatively"). This progressive semantic transition convincingly demonstrates that the extended trajectories within the Complex dataset encapsulate genuine trial-and-error, hypothesis exploration, and self-correction behaviors, which are the quintessential elements of advanced Long-CoT reasoning.

\section{Extended Related Work}
\paragraph{Learning from Negative Samples.}
Utilizing incorrect data to enhance the reasoning capabilities of LLMs has gained significant attention, primarily through in-context learning and optimization-based approaches.
Instead of relying solely on high-quality demonstrations, recent studies show that injecting flawed examples directly into the prompt can help models delineate task boundaries~\citep{alazraki2025no}. Alternatively, optimization-based methods actively update model weights to penalize or refine erroneous outputs during training.
This is typically achieved by constructing positive-negative pairs to maximize the likelihood margin in preference learning~\citep{rafailov2023direct,guo2025deepseek,lu2026one}, training models on flawed trajectories to trigger iterative self-correction~\citep{an2023learning}, or explicitly minimizing the likelihood of negative samples to encourage broader exploration and significantly improve \textit{pass@k} performance~\citep{zhusurprising}.
While existing literature predominantly treats negative samples strictly as penalization targets or examples of what to avoid, our findings fundamentally challenge this paradigm. We demonstrate that factually flawed trajectories can actually serve as highly effective positive supervision for Long-CoT distillation, provided their structural reasoning behaviors remain intact.

\section{Additional Experiments Following Peer Review}
\label{app:rebuttal}
We report further controls on training domain, teacher choice, more recent test sets, and hidden-state representations. P and N denote trajectories selected by correct and incorrect final answers, respectively; mathematical averages are computed over the five benchmarks in \Cref{sec:pvsn}.

\paragraph{Science-domain supervision.}
We fine-tune Qwen3 base models on correct (SFT-P) or incorrect (SFT-N) trajectories from a science QA subset of GURU, then evaluate both mathematical and non-mathematical reasoning. \Cref{tab:guru_transfer} shows similar improvements from both groups of trajectories, including when assessed on mathematical problems outside the science training domain. Non-math average is the unweighted mean over SuperGLUE, GPQA-Diamond, and MMLU-Pro.
\begin{table}[htbp]
\centering
\small
\caption{Cross-domain evaluation after science QA (GURU) training. The base models are evaluated without fine-tuning; P/N distinguish final-answer correctness of the training trajectories. Values are percentages.}
\label{tab:guru_transfer}
\resizebox{0.94\textwidth}{!}{%
\begin{tabular}{llrrrrr}
\toprule
Student & Training & Math avg. & SuperGLUE & GPQA-D & MMLU-Pro & Non-math avg. \\
\midrule
\multirow{3}{*}{Qwen3-1.7B-Base} & Base & 6.9 & 10.6 & 15.9 & 27.3 & 17.9 \\
 & SFT-P & 27.2 & 18.0 & 22.0 & 34.9 & 25.0 \\
 & SFT-N & 26.8 & 18.3 & 24.5 & 34.0 & 25.6 \\
\midrule
\multirow{3}{*}{Qwen3-4B-Base} & Base & 19.1 & 8.5 & 16.0 & 45.6 & 23.4 \\
 & SFT-P & 44.6 & 29.1 & 34.8 & 49.0 & 37.6 \\
 & SFT-N & 43.7 & 29.8 & 36.4 & 49.4 & 38.5 \\
\bottomrule
\end{tabular}}
\end{table}

\paragraph{Alternative teachers and thinking-mode traces.}
The main study samples from Qwen3-4B in non-thinking mode. We repeat the correct-versus-incorrect comparison using Gemma3-12B-it as a different teacher family and Qwen3-4B in thinking mode to sample extended CoTs. Both are evaluated on Qwen3-1.7B-Base and Qwen3-4B-Base with the same five math benchmarks. As summarized in \Cref{tab:teacher_robustness}, training on incorrect final answers remains competitive across these settings.
\begin{table}[htbp]
\centering
\small
\caption{Teacher and sampling-mode controls: avg@16 accuracy (\%) across the five mathematical benchmarks. Base is the corresponding student before fine-tuning.}
\label{tab:teacher_robustness}
\begin{tabular}{llrrr}
\toprule
Teacher / mode & Student & Base & SFT-P & SFT-N \\
\midrule
\multirow{2}{*}{Gemma3-12B-it} & Qwen3-1.7B-Base & 6.9 & 22.8 & 22.2 \\
 & Qwen3-4B-Base & 19.1 & 38.5 & 38.4 \\
\midrule
\multirow{2}{*}{Qwen3-4B / thinking} & Qwen3-1.7B-Base & 6.9 & 20.9 & 20.4 \\
 & Qwen3-4B-Base & 19.1 & 46.5 & 44.4 \\
\bottomrule
\end{tabular}
\end{table}

\paragraph{More recent benchmarks.}
We evaluate the four student architectures in \Cref{fig:Pos_vs_Neg} on AIME 2026 and LiveCodeBench-v6. \Cref{tab:new_benchmarks} averages the accuracy of Qwen3-1.7B/4B-Base, Qwen2.5-7B, and Llama-3.1-8B-Instruct for each condition. Flawed-trajectory SFT remains comparable to correct-trajectory SFT with either objective, as in the original benchmarks.
\begin{table}[htbp]
\centering
\small
\caption{Correct vs. incorrect trajectory distillation on newer evaluation sets; percentages averaged over four student architectures (rounded to one decimal place).}
\label{tab:new_benchmarks}
\begin{tabular}{llrr}
\toprule
Loss & Training data & AIME 2026 & LiveCodeBench-v6 \\
\midrule
-- & Base & 1.3 & 9.3 \\
\multirow{2}{*}{$-\log p$} & P & 5.5 & 16.6 \\
 & N & 5.6 & 16.9 \\
\multirow{2}{*}{$-p$} & P & 6.2 & 16.0 \\
 & N & 7.1 & 16.0 \\
\bottomrule
\end{tabular}
\end{table}

\paragraph{Representation-level comparison.}
We compute linear centered kernel alignment (CKA)~\citep{kornblith2019similarity} between token-aligned late-layer response representations of Qwen3-4B variants, using 128 held-out MATH problems paired with identical reference reasoning traces. \Cref{tab:cka} reports strong agreement between models trained on correct and incorrect trajectories, compared with their similarity to the base model and to a model trained on NuminaMath-CoT trajectories. Shared evaluation inputs keep the input traces identical across model pairs.
\begin{table}[htbp]
\centering
\small
\caption{Linear CKA between late-layer response-token representations on the same held-out MATH inputs and reference traces (128 problems). SFT-numina is fine-tuned on NuminaMath-CoT.}
\label{tab:cka}
\begin{tabular}{lr}
\toprule
Qwen3-4B model pair & Linear CKA \\
\midrule
Base vs. SFT-P & 0.383 \\
Base vs. SFT-N & 0.397 \\
SFT-P vs. SFT-N & 0.998 \\
SFT-P vs. SFT-numina & 0.412 \\
\bottomrule
\end{tabular}
\end{table}
The near-identical representations learned from positive and negative traces, despite their divergence from the base model, are consistent with a shared reasoning pattern at the representation level.

\section{Detailed Experimental Results}

\begin{table}[ht]
\centering
\small
\caption{Detailed avg@16 Evaluation Results Corresponding to \Cref{fig:Pos_vs_Neg}.}
\resizebox{0.95\textwidth}{!}{
\setlength{\tabcolsep}{5pt}
\begin{tabular}{llccccccl}
\toprule
\textbf{Model} & \textbf{Method} & \textbf{Data} & \textbf{AIME 2024} & \textbf{AMC23} & \textbf{Minerva Math} & \textbf{Olympiad Bench} & \textbf{MATH} & \textbf{Average} \\ \midrule

\cellcolor{white} & Base & - & 1.3 & 6.0 & 5.0 & 6.2 & 16.0 & 6.9 \\
\rowcolor{gcolor1} \cellcolor{white} & -logp & P & 4.4 & 29.1 & 21.5 & 20.9 & 56.8 & 26.5 \\
\rowcolor{gcolor1} \cellcolor{white} & -logp & N & 2.1 & 31.9 & 20.2 & 20.9 & 54.2 & 25.9 \\
\rowcolor{gcolor2} \cellcolor{white} & -p & P & 7.3 & 35.6 & 24.1 & 24.7 & 61.7 & 30.7 \\
\rowcolor{gcolor2} \cellcolor{white}\multirow{-5}{*}{\textbf{Qwen3-1.7B-Base}} & -p & N & 7.7 & 34.8 & 24.5 & 24.0 & 61.3 & 30.5 \\ \midrule

\cellcolor{white} & Base & - & 5.8 & 23.8 & 11.5 & 18.2 & 36.1 & 19.1 \\
\rowcolor{gcolor1} \cellcolor{white} & -logp & P & 12.3 & 51.4 & 34.5 & 39.2 & 76.4 & 42.8 \\
\rowcolor{gcolor1} \cellcolor{white} & -logp & N & 15.0 & 52.0 & 33.0 & 39.7 & 76.4 & 43.2 \\
\rowcolor{gcolor2} \cellcolor{white} & -p & P & 15.6 & 53.1 & 33.5 & 39.1 & 79.3 & 44.1 \\
\rowcolor{gcolor2} \cellcolor{white}\multirow{-5}{*}{\textbf{Qwen3-4B-Base}} & -p & N & 15.2 & 49.4 & 32.3 & 38.7 & 77.8 & 42.7 \\ \midrule

\cellcolor{white} & Base & - & 2.5 & 24.2 & 14.1 & 16.0 & 43.8 & 20.1 \\
\rowcolor{gcolor1} \cellcolor{white} & -logp & P & 12.1 & 50.0 & 28.9 & 33.7 & 71.5 & 39.2 \\
\rowcolor{gcolor1} \cellcolor{white} & -logp & N & 10.6 & 45.5 & 28.8 & 34.1 & 71.1 & 38.0 \\
\rowcolor{gcolor2} \cellcolor{white} & -p & P & 13.7 & 47.5 & 31.4 & 33.3 & 72.0 & 39.6 \\
\rowcolor{gcolor2} \cellcolor{white}\multirow{-5}{*}{\textbf{Qwen2.5-7B}} & -p & N & 13.7 & 43.8 & 29.8 & 33.4 & 73.0 & 38.7 \\ \midrule

\cellcolor{white} & Base & - & 1.9 & 16.1 & 18.5 & 10.8 & 40.4 & 17.5 \\
\rowcolor{gcolor1} \cellcolor{white} & -logp & P & 3.3 & 33.8 & 26.5 & 22.4 & 59.8 & 29.2 \\
\rowcolor{gcolor1} \cellcolor{white} & -logp & N & 4.4 & 29.1 & 24.7 & 20.5 & 54.6 & 26.7 \\
\rowcolor{gcolor2} \cellcolor{white} & -p & P & 3.9 & 36.1 & 29.6 & 22.8 & 55.7 & 29.6 \\
\rowcolor{gcolor2} \cellcolor{white}\multirow{-5}{*}{\textbf{Llama3.1-8B-Instruct}} & -p & N & 6.9 & 32.0 & 26.6 & 22.8 & 56.7 & 29.0 \\ \bottomrule

\end{tabular}
}
\end{table}

\begin{table}[ht]
\centering
\small
\caption{Detailed pass@16 Evaluation Results Corresponding to \Cref{fig:Pos_vs_Neg}.}
\resizebox{0.95\textwidth}{!}{
\setlength{\tabcolsep}{5pt}
\begin{tabular}{llccccccl}
\toprule
\textbf{Model} & \textbf{Method} & \textbf{Data} & \textbf{AIME 2024} & \textbf{AMC23} & \textbf{Minerva Math} & \textbf{Olympiad Bench} & \textbf{MATH} & \textbf{Average} \\ \midrule

\cellcolor{white} & Base & - & 10.0 & 75.0 & 39.3 & 47.7 & 80.4 & 50.5 \\
\rowcolor{gcolor1} \cellcolor{white} & -logp & P & 20.0 & 75.0 & 55.1 & 51.0 & 86.3 & 57.5 \\
\rowcolor{gcolor1} \cellcolor{white} & -logp & N & 13.3 & 77.5 & 52.9 & 49.3 & 87.3 & 56.1  \\
\rowcolor{gcolor2} \cellcolor{white} & -p & P & 20.0 & 77.5 & 49.6 & 49.9 & 86.1 & 56.6 \\
\rowcolor{gcolor2} \cellcolor{white}\multirow{-5}{*}{\textbf{Qwen3-1.7B-Base}} & -p & N & 16.7 & 82.5 & 51.1 & 48.1 & 84.3 & 56.5  \\ \midrule

\cellcolor{white} & Base & - & 30.0 & 87.5 & 49.6 & 58.7 & 88.2 & 62.8 \\
\rowcolor{gcolor1} \cellcolor{white} & -logp & P & 30.0 & 95.0 & 63.2 & 66.1 & 93.7 & 69.6 \\
\rowcolor{gcolor1} \cellcolor{white} & -logp & N & 36.7 & 87.5 & 63.6 & 66.2 & 93.7 & 69.5 \\
\rowcolor{gcolor2} \cellcolor{white} & -p & P & 26.7 & 80.0 & 53.7 & 59.6 & 92.5 & 62.5 \\
\rowcolor{gcolor2} \cellcolor{white}\multirow{-5}{*}{\textbf{Qwen3-4B-Base}} & -p & N & 36.7 & 85.0 & 53.3 & 60.0 & 92.5 & 65.5 \\ \midrule

\cellcolor{white} & Base & - & 23.3 & 70.0 & 50.7 & 51.9 & 85.4 & 56.3 \\
\rowcolor{gcolor1} \cellcolor{white} & -logp & P & 43.3 & 80.0 & 60.7 & 64.1 & 92.5 & 68.1 \\
\rowcolor{gcolor1} \cellcolor{white} & -logp & N & 30.0 & 87.5 & 61.0 & 65.0 & 92.3 & 67.2 \\
\rowcolor{gcolor2} \cellcolor{white} & -p & P & 23.3 & 82.5 & 53.3 & 55.6 & 90.3 & 61.0 \\
\rowcolor{gcolor2} \cellcolor{white}\multirow{-5}{*}{\textbf{Qwen2.5-7B}} & -p & N & 26.7 & 77.5 & 56.2 & 56.3 & 89.9 & 61.3 \\ \midrule

\cellcolor{white} & Base & - & 16.7 & 60.0 & 51.1 & 40.3 & 78.6 & 49.3 \\
\rowcolor{gcolor1} \cellcolor{white} & -logp & P & 20.0 & 80.0 & 61.8 & 51.9 & 89.9 & 60.7 \\
\rowcolor{gcolor1} \cellcolor{white} & -logp & N & 26.7 & 75.0 & 59.6 & 52.4 & 85.3 & 59.8 \\
\rowcolor{gcolor2} \cellcolor{white} & -p & P & 23.3 & 85.0 & 62.5 & 52.3 & 85.6 & 61.7 \\
\rowcolor{gcolor2} \cellcolor{white}\multirow{-5}{*}{\textbf{Llama3.1-8B-Instruct}} & -p & N & 33.3 & 87.5 & 57.4 & 50.7 & 85.0 & 62.8 \\ \bottomrule

\end{tabular}
}
\end{table}

\begin{table}[ht]
\centering
\small
\caption{Averaged Response Token Lengths Corresponding to \Cref{fig:Pos_vs_Neg}.}
\resizebox{0.95\textwidth}{!}{
\setlength{\tabcolsep}{5pt}
\begin{tabular}{llccccccl}
\toprule
\textbf{Model} & \textbf{Method} & \textbf{Data} & \textbf{AIME 2024} & \textbf{AMC23} & \textbf{Minerva Math} & \textbf{Olympiad Bench} & \textbf{MATH} & \textbf{Average} \\ \midrule

\cellcolor{white} & Base & - & 1123.7 & 989.6 & 765.3 & 916.7 & 747.3 & 908.5 \\
\rowcolor{gcolor1} \cellcolor{white} & -logp & P & 2419.3 & 1686.8 & 905.7 & 1818 & 1002.5 & 1566.5 \\
\rowcolor{gcolor1} \cellcolor{white} & -logp & N & 2510.3 & 1762.9 & 980.4 & 1802.3 & 1066.2 & 1624.4 \\
\rowcolor{gcolor2} \cellcolor{white} & -p & P & 3172.3 & 2115 & 998.8 & 2323 & 1161.8 & 1954.2 \\
\rowcolor{gcolor2} \cellcolor{white}\multirow{-5}{*}{\textbf{Qwen3-1.7B-Base}} & -p & N & 3318.5 & 2210.7 & 1092.8 & 2466.4 & 1247.7 & 2067.2  \\ \midrule

\cellcolor{white} & Base & - & 1532.7 & 1312.2 & 1451.7 & 1348.7 & 1266.9 & 1382.4 \\
\rowcolor{gcolor1} \cellcolor{white} & -logp & P & 2575.3 & 1650.3 & 756.6 & 1623.8 & 878.3 & 1496.9 \\
\rowcolor{gcolor1} \cellcolor{white} & -logp & N & 2514.1 & 1671.5 & 799.3 & 1597.9 & 890 & 1494.6  \\
\rowcolor{gcolor2} \cellcolor{white} & -p & P & 3117.7 & 1869 & 772.7 & 1946.3 & 910.9 & 1723.3 \\
\rowcolor{gcolor2} \cellcolor{white}\multirow{-5}{*}{\textbf{Qwen3-4B-Base}} & -p & N & 3192.7 & 1873.7 & 830.2 & 1996 & 970.2 & 1772.6  \\ \midrule

\cellcolor{white} & Base & - & 822.9 & 748 & 699.6 & 741.4 & 589.6 & 720.3 \\
\rowcolor{gcolor1} \cellcolor{white} & -logp & P & 2498.1 & 1644.5 & 932 & 1688.6 & 919.4 & 1536.5 \\
\rowcolor{gcolor1} \cellcolor{white} & -logp & N & 2560.5 & 1710.8 & 973.6 & 1711.7 & 958.3 & 1583.0  \\
\rowcolor{gcolor2} \cellcolor{white} & -p & P & 3095.7 & 1827.1 & 994.6 & 2131.3 & 1041.2 & 1818.0 \\
\rowcolor{gcolor2} \cellcolor{white}\multirow{-5}{*}{\textbf{Qwen2.5-7B}} & -p & N & 3304.2 & 2047.9 & 1102.1 & 2180.5 & 1101 & 1947.1 \\ \midrule

\cellcolor{white} & Base & - & 1376.4 & 1204.9 & 784.3 & 1267.1 & 740.8 & 1074.7 \\
\rowcolor{gcolor1} \cellcolor{white} & -logp & P & 2446.5 & 1666.4 & 859.1 & 1784.2 & 923.5 & 1535.9 \\
\rowcolor{gcolor1} \cellcolor{white} & -logp & N & 2397.4 & 1706.9 & 872.1 & 1749.6 & 952.2 & 1535.6  \\
\rowcolor{gcolor2} \cellcolor{white} & -p & P & 2486 & 1639.5 & 803.6 & 1930 & 1021.2 & 1576.1 \\
\rowcolor{gcolor2} \cellcolor{white}\multirow{-5}{*}{\textbf{Llama3.1-8B-Instruct}} & -p & N & 2813.4 & 1848.6 & 913.6 & 2105.2 & 1101.3 & 1756.4  \\ \bottomrule

\end{tabular}
}
\end{table}

\begin{table}[ht]
\centering
\small
\caption{Averaged Response Token Lengths Corresponding to \Cref{fig:loss_complexity}.}
\resizebox{0.95\textwidth}{!}{
\setlength{\tabcolsep}{5pt}
\begin{tabular}{lllcccccc}
\toprule
\textbf{Model} & \textbf{Dataset} & \textbf{Loss} & \textbf{AIME 2024} & \textbf{AMC23} & \textbf{Minerva Math} & \textbf{Olympiad Bench} & \textbf{MATH} & \textbf{Average} \\ \midrule

\rowcolor{gcolor1} \cellcolor{white} & \cellcolor{white} & -logp & 763.3 & 639 & 643.7 & 677.7 & 442 & 633.1 \\
\rowcolor{gcolor2} \cellcolor{white} & \cellcolor{white}\multirow{-2}{*}{Simple} & -p & 1400.8 & 913.6 & 684.9 & 956.8 & 573.2 & 905.9 \\
\rowcolor{gcolor1} \cellcolor{white} & \cellcolor{white} & -logp & 2575.3 & 1650.3 & 756.6 & 1623.8 & 878.3 & 1496.9 \\
\rowcolor{gcolor2} \cellcolor{white} & \cellcolor{white}\multirow{-2}{*}{Medium} & -p & 3117.7 & 1869 & 772.7 & 1946.3 & 910.9 & 1723.3 \\
\rowcolor{gcolor1} \cellcolor{white} & \cellcolor{white} & -logp & 19348.9 & 12440.8 & 8881.1 & 12102.3 & 8281.8 & 12211.0 \\
\rowcolor{gcolor2} \cellcolor{white}\multirow{-6}{*}{\textbf{Qwen3-4B-Base}} & \cellcolor{white}\multirow{-2}{*}{Complex} & -p & 12524.4 & 6254.3 & 2616.5 & 6253.2 & 3230.8 & 6175.8 \\ \midrule

\rowcolor{gcolor1} \cellcolor{white} & \cellcolor{white} & -logp & 737.2 & 649.6 & 514.8 & 679.3 & 460.2 & 608.2 \\
\rowcolor{gcolor2} \cellcolor{white} & \cellcolor{white}\multirow{-2}{*}{Simple} & -p & 1310.4 & 794.1 & 547.1 & 37.35 & 525.9 & 643.0 \\
\rowcolor{gcolor1} \cellcolor{white} & \cellcolor{white} & -logp & 2378.8 & 1462.3 & 674.4 & 1506.9 & 780 & 1360.5 \\
\rowcolor{gcolor2} \cellcolor{white} & \cellcolor{white}\multirow{-2}{*}{Medium} & -p & 3176.2 & 1729 & 719.8 & 1896.4 & 885.8 & 1681.4 \\
\rowcolor{gcolor1} \cellcolor{white} & \cellcolor{white} & -logp & 12705.1 & 7637.9 & 5366.7 & 7703.5 & 4663.7 & 7615.4 \\
\rowcolor{gcolor2} \cellcolor{white}\multirow{-6}{*}{\textbf{Qwen3-14B-Base}} & \cellcolor{white}\multirow{-2}{*}{Complex} & -p & 14752.5 & 7494.1 & 1456.3 & 7337.4 & 2962.3 & 6800.5 \\ \midrule

\rowcolor{gcolor1} \cellcolor{white} & \cellcolor{white} & -logp & 628.8 & 535.6 & 597.9 & 644.0 & 427.6 & 547.5 \\
\rowcolor{gcolor2} \cellcolor{white} & \cellcolor{white}\multirow{-2}{*}{Simple} & -p & 1029.3 & 701.9 & 564.4 & 812.3 & 468.6 & 715.3 \\
\rowcolor{gcolor1} \cellcolor{white} & \cellcolor{white} & -logp & 2498.1 & 1644.5 & 932 & 1688.6 & 919.4 & 1536.5 \\
\rowcolor{gcolor2} \cellcolor{white} & \cellcolor{white}\multirow{-2}{*}{Medium} & -p & 3095.7 & 1827.1 & 994.6 & 2131.3 & 1041.2 & 1818.0 \\
\rowcolor{gcolor1} \cellcolor{white} & \cellcolor{white} & -logp & 22299.1 & 14533.1 & 10226.6 & 14354.1 & 7673 & 13817.2 \\
\rowcolor{gcolor2} \cellcolor{white}\multirow{-6}{*}{\textbf{Qwen2.5-7B}} & \cellcolor{white}\multirow{-2}{*}{Complex} & -p & 32269.2 & 31560.7 & 20972.8 & 31208.2 & 29950.4 & 29192.3 \\ \midrule

\rowcolor{gcolor1} \cellcolor{white} & \cellcolor{white} & -logp & 777.1 & 663 & 595.8 & 734.4 & 486.4 & 651.3 \\
\rowcolor{gcolor2} \cellcolor{white} & \cellcolor{white}\multirow{-2}{*}{Simple} & -p & 889.2 & 697.3 & 502.4 & 731.1 & 439.4 & 651.9 \\
\rowcolor{gcolor1} \cellcolor{white} & \cellcolor{white} & -logp & 2446.5 & 1666.4 & 859.1 & 1784.2 & 923.5 & 1535.9 \\
\rowcolor{gcolor2} \cellcolor{white} & \cellcolor{white}\multirow{-2}{*}{Medium} & -p & 2486 & 1639.5 & 803.6 & 1930 & 1021.2 & 1576.1 \\
\rowcolor{gcolor1} \cellcolor{white} & \cellcolor{white} & -logp & 19768.9 & 14155.6 & 6962.8 & 13722.8 & 8127.2 & 12547.5 \\
\rowcolor{gcolor2} \cellcolor{white}\multirow{-6}{*}{\textbf{Llama3.1-8B-Instruct}} & \cellcolor{white}\multirow{-2}{*}{Complex} & -p & 30295.9 & 23400.8 & 13187 & 22117.1 & 15326.5 & 20865.5 \\ \bottomrule

\end{tabular}
}
\end{table}

\begin{table}[ht]
\centering
\small
\caption{Averaged Response Token Lengths Corresponding to \Cref{fig:loss_complexity}.}
\label{tab:loss_length}
\resizebox{0.95\textwidth}{!}{
\setlength{\tabcolsep}{5pt}
\begin{tabular}{lllcccccc}
\toprule
\textbf{Model} & \textbf{Dataset} & \textbf{Loss} & \textbf{AIME 2024} & \textbf{AMC23} & \textbf{Minerva Math} & \textbf{Olympiad Bench} & \textbf{MATH} & \textbf{Average} \\ \midrule

\rowcolor{gcolor1} \cellcolor{white} & \cellcolor{white} & -logp & 763.3 & 639 & 643.7 & 677.7 & 442 & 633.1 \\
\rowcolor{gcolor2} \cellcolor{white} & \cellcolor{white}\multirow{-2}{*}{Simple} & -p & 1400.8 & 913.6 & 684.9 & 956.8 & 573.2 & 905.9 \\
\rowcolor{gcolor1} \cellcolor{white} & \cellcolor{white} & -logp & 2575.3 & 1650.3 & 756.6 & 1623.8 & 878.3 & 1496.9 \\
\rowcolor{gcolor2} \cellcolor{white} & \cellcolor{white}\multirow{-2}{*}{Medium} & -p & 3117.7 & 1869 & 772.7 & 1946.3 & 910.9 & 1723.3 \\
\rowcolor{gcolor1} \cellcolor{white} & \cellcolor{white} & -logp & 19348.9 & 12440.8 & 8881.1 & 12102.3 & 8281.8 & 12211.0 \\
\rowcolor{gcolor2} \cellcolor{white}\multirow{-6}{*}{\textbf{Qwen3-4B-Base}} & \cellcolor{white}\multirow{-2}{*}{Complex} & -p & 12524.4 & 6254.3 & 2616.5 & 6253.2 & 3230.8 & 6175.8 \\ \midrule

\rowcolor{gcolor1} \cellcolor{white} & \cellcolor{white} & -logp & 737.2 & 649.6 & 514.8 & 679.3 & 460.2 & 608.2 \\
\rowcolor{gcolor2} \cellcolor{white} & \cellcolor{white}\multirow{-2}{*}{Simple} & -p & 1310.4 & 794.1 & 547.1 & 37.35 & 525.9 & 643.0 \\
\rowcolor{gcolor1} \cellcolor{white} & \cellcolor{white} & -logp & 2378.8 & 1462.3 & 674.4 & 1506.9 & 780 & 1360.5 \\
\rowcolor{gcolor2} \cellcolor{white} & \cellcolor{white}\multirow{-2}{*}{Medium} & -p & 3176.2 & 1729 & 719.8 & 1896.4 & 885.8 & 1681.4 \\
\rowcolor{gcolor1} \cellcolor{white} & \cellcolor{white} & -logp & 12705.1 & 7637.9 & 5366.7 & 7703.5 & 4663.7 & 7615.4 \\
\rowcolor{gcolor2} \cellcolor{white}\multirow{-6}{*}{\textbf{Qwen3-14B-Base}} & \cellcolor{white}\multirow{-2}{*}{Complex} & -p & 14752.5 & 7494.1 & 1456.3 & 7337.4 & 2962.3 & 6800.5 \\ \midrule

\rowcolor{gcolor1} \cellcolor{white} & \cellcolor{white} & -logp & 628.8 & 535.6 & 597.9 & 644.0 & 427.6 & 547.5 \\
\rowcolor{gcolor2} \cellcolor{white} & \cellcolor{white}\multirow{-2}{*}{Simple} & -p & 1029.3 & 701.9 & 564.4 & 812.3 & 468.6 & 715.3 \\
\rowcolor{gcolor1} \cellcolor{white} & \cellcolor{white} & -logp & 2498.1 & 1644.5 & 932 & 1688.6 & 919.4 & 1536.5 \\
\rowcolor{gcolor2} \cellcolor{white} & \cellcolor{white}\multirow{-2}{*}{Medium} & -p & 3095.7 & 1827.1 & 994.6 & 2131.3 & 1041.2 & 1818.0 \\
\rowcolor{gcolor1} \cellcolor{white} & \cellcolor{white} & -logp & 22299.1 & 14533.1 & 10226.6 & 14354.1 & 7673 & 13817.2 \\
\rowcolor{gcolor2} \cellcolor{white}\multirow{-6}{*}{\textbf{Qwen2.5-7B}} & \cellcolor{white}\multirow{-2}{*}{Complex} & -p & 32269.2 & 31560.7 & 20972.8 & 31208.2 & 29950.4 & 29192.3 \\ \midrule

\rowcolor{gcolor1} \cellcolor{white} & \cellcolor{white} & -logp & 777.1 & 663 & 595.8 & 734.4 & 486.4 & 651.3 \\
\rowcolor{gcolor2} \cellcolor{white} & \cellcolor{white}\multirow{-2}{*}{Simple} & -p & 889.2 & 697.3 & 502.4 & 731.1 & 439.4 & 651.9 \\
\rowcolor{gcolor1} \cellcolor{white} & \cellcolor{white} & -logp & 2446.5 & 1666.4 & 859.1 & 1784.2 & 923.5 & 1535.9 \\
\rowcolor{gcolor2} \cellcolor{white} & \cellcolor{white}\multirow{-2}{*}{Medium} & -p & 2486 & 1639.5 & 803.6 & 1930 & 1021.2 & 1576.1 \\
\rowcolor{gcolor1} \cellcolor{white} & \cellcolor{white} & -logp & 19768.9 & 14155.6 & 6962.8 & 13722.8 & 8127.2 & 12547.5 \\
\rowcolor{gcolor2} \cellcolor{white}\multirow{-6}{*}{\textbf{Llama3.1-8B-Instruct}} & \cellcolor{white}\multirow{-2}{*}{Complex} & -p & 30295.9 & 23400.8 & 13187 & 22117.1 & 15326.5 & 20865.5 \\ \bottomrule

\end{tabular}
}
\end{table}

\begin{table}[ht]
\centering
\small
\caption{The detailed accuracy in \Cref{fig:loss_complexity}.}
\resizebox{0.95\textwidth}{!}{
\setlength{\tabcolsep}{5pt}
\begin{tabular}{lllcccccl}
\toprule
\textbf{Model} & \textbf{CoT} & \textbf{Loss} & \textbf{AIME 2024} & \textbf{AMC23} & \textbf{Minerva Math} & \textbf{Olympiad Bench} & \textbf{MATH} & \textbf{Average} \\ \midrule

\rowcolor{gcolor1} \cellcolor{white} & \cellcolor{white} & -logp & 3.3 & 25.9 & 17.8 & 17.4 & 49.7 & \textbf{22.8} \\
\rowcolor{gcolor2} \cellcolor{white} & \cellcolor{white}\multirow{-2}{*}{Simple} & -p & 8.8 & 43.0 & 29.6 & 33.2 & 72.2 & \textbf{37.4} \posdiff{14.6} \\
\rowcolor{gcolor1} \cellcolor{white} & \cellcolor{white} & -logp & 12.3 & 51.4 & 34.5 & 39.2 & 76.4 & \textbf{42.8} \\
\rowcolor{gcolor2} \cellcolor{white} & \cellcolor{white}\multirow{-2}{*}{Medium} & -p & 15.6 & 53.1 & 33.5 & 39.1 & 79.3 & \textbf{44.1} \posdiff{1.3} \\
\rowcolor{gcolor1} \cellcolor{white} & \cellcolor{white} & -logp & 24.4 & 57.3 & 30.0 & 44.4 & 71.9 & \textbf{45.6} \\
\rowcolor{gcolor2} \cellcolor{white}\multirow{-6}{*}{\textbf{Qwen3-4B-Base}} & \cellcolor{white}\multirow{-2}{*}{Complex} & -p & 12.5 & 47.2 & 31.7 & 36.9 & 74.2 & \textbf{40.5} \negdiff{5.1} \\ \midrule

\rowcolor{gcolor1} \cellcolor{white} & \cellcolor{white} & -logp & 2.9 & 33.6 & 26.2 & 22.6 & 61.1 & \textbf{29.3} \\
\rowcolor{gcolor2} \cellcolor{white} & \cellcolor{white}\multirow{-2}{*}{Simple} & -p & 10.2 & 52.0 & 39.3 & 37.4 & 76.3 & \textbf{43.0} \posdiff{13.7} \\
\rowcolor{gcolor1} \cellcolor{white} & \cellcolor{white} & -logp & 17.5 & 57.3 & 39.1 & 42.4 & 81.2 & \textbf{47.5} \\
\rowcolor{gcolor2} \cellcolor{white} & \cellcolor{white}\multirow{-2}{*}{Medium} & -p & 15.4 & 60.5 & 40.4 & 42.9 & 83.0 & \textbf{48.4} \posdiff{0.9} \\
\rowcolor{gcolor1} \cellcolor{white} & \cellcolor{white} & -logp & 33.1 & 71.1 & 43.3 & 50.2 & 82.7 & \textbf{56.1} \\
\rowcolor{gcolor2} \cellcolor{white}\multirow{-6}{*}{\textbf{Qwen3-14B-Base}} & \cellcolor{white}\multirow{-2}{*}{Complex} & -p & 14.6 & 60.8 & 39.5 & 42.5 & 82.7 & \textbf{48.0} \negdiff{8.1} \\ \midrule

\rowcolor{gcolor1} \cellcolor{white} & \cellcolor{white} & -logp & 1.7 & 22.8 & 14.2 & 14.6 & 46.0 & \textbf{19.9} \\
\rowcolor{gcolor2} \cellcolor{white} & \cellcolor{white}\multirow{-2}{*}{Simple} & -p & 6.1 & 39.4 & 28.8 & 29.4 & 65.6 & \textbf{33.9} \posdiff{14.0} \\
\rowcolor{gcolor1} \cellcolor{white} & \cellcolor{white} & -logp & 12.1 & 50.0 & 28.9 & 33.7 & 71.5 & \textbf{39.2} \\
\rowcolor{gcolor2} \cellcolor{white} & \cellcolor{white}\multirow{-2}{*}{Medium} & -p & 13.7 & 47.5 & 31.4 & 33.3 & 72.0 & \textbf{39.6} \posdiff{0.4} \\
\rowcolor{gcolor1} \cellcolor{white} & \cellcolor{white} & -logp & 17.7 & 56.6 & 36.3 & 42.3 & 79.3 & \textbf{46.4} \\
\rowcolor{gcolor2} \cellcolor{white}\multirow{-6}{*}{\textbf{Qwen2.5-7B}} & \cellcolor{white}\multirow{-2}{*}{Complex} & -p & 2.7 & 15.8 & 22.9 & 13.4 & 30.2 & \textbf{17.0} \negdiff{29.4} \\ \midrule

\rowcolor{gcolor1} \cellcolor{white} & \cellcolor{white} & -logp & 0.2 & 11.7 & 13.0 & 8.5 & 33.3 & \textbf{13.3} \\
\rowcolor{gcolor2} \cellcolor{white} & \cellcolor{white}\multirow{-2}{*}{Simple} & -p & 3.7 & 21.1 & 23.4 & 12.8 & 48.0 & \textbf{21.8} \posdiff{8.5} \\
\rowcolor{gcolor1} \cellcolor{white} & \cellcolor{white} & -logp & 3.3 & 33.8 & 26.5 & 22.4 & 59.8 & \textbf{29.2} \\
\rowcolor{gcolor2} \cellcolor{white} & \cellcolor{white}\multirow{-2}{*}{Medium} & -p & 3.9 & 36.1 & 29.6 & 22.8 & 55.7 & \textbf{29.6} \posdiff{0.4} \\
\rowcolor{gcolor1} \cellcolor{white} & \cellcolor{white} & -logp & 5.4 & 37.0 & 25.8 & 25.0 & 58.2 & \textbf{30.3} \\
\rowcolor{gcolor2} \cellcolor{white}\multirow{-6}{*}{\textbf{Llama3.1-8B}} & \cellcolor{white}\multirow{-2}{*}{Complex} & -p & 3.1 & 24.8 & 21.9 & 18.6 & 48.0 & \textbf{23.3} \negdiff{7.0} \\ \bottomrule

\end{tabular}
}
\end{table}

\begin{table}[ht]
\centering
\small
\caption{Detailed avg@16 Evaluation Results Corresponding to \Cref{fig:prompt_diversity}.}
\label{tab:prompt_diversity_fixed_budget}
\resizebox{0.95\textwidth}{!}{
\setlength{\tabcolsep}{5pt}
\begin{tabular}{llcccccc}
\toprule
\textbf{Model} & \textbf{Setting} & \textbf{AIME 2024} & \textbf{AMC23} & \textbf{Minerva Math} & \textbf{Olympiad Bench} & \textbf{MATH} & \textbf{Average} \\ \midrule

 & $500 \times 128$ & 5.2 & 30.9 & 22.3 & 22.6 & 58.8 & 28.0 \\
 & $1\text{k} \times 64$ & 3.5 & 30.2 & 21.9 & 22.5 & 59.2 & 27.5 \\
 & $2\text{k} \times 32$ & 3.3 & 32.2 & 21.4 & 21.9 & 56.6 & 27.1 \\
 & $4\text{k} \times 16$ & 4.2 & 30.9 & 21.3 & 21.9 & 56.6 & 27.0 \\
 & $8\text{k} \times 8$ & 5.6 & 28.6 & 20.7 & 21.6 & 56.0 & 26.5 \\
\multirow{-6}{*}{\textbf{Qwen3-1.7B-Base}} & $16\text{k} \times 4$ & 4.0 & 30.5 & 21.2 & 21.3 & 57.2 & 26.8 \\ \midrule

 & $500 \times 128$ & 12.7 & 52.3 & 35.5 & 40.4 & 78.2 & 43.8 \\
 & $1\text{k} \times 64$ & 14.6 & 54.4 & 34.0 & 40.3 & 78.6 & 44.4 \\
 & $2\text{k} \times 32$ & 15.2 & 54.7 & 34.6 & 40.2 & 77.4 & 44.4 \\
 & $4\text{k} \times 16$ & 15.6 & 51.6 & 34.2 & 39.5 & 79.7 & 44.1 \\
 & $8\text{k} \times 8$ & 16.3 & 53.6 & 34.5 & 39.9 & 77.3 & 44.3 \\
\multirow{-6}{*}{\textbf{Qwen3-4B-Base}} & $16\text{k} \times 4$ & 14.0 & 53.0 & 34.6 & 39.4 & 78.1 & 43.8 \\ \midrule

 & $500 \times 128$ & 9.6 & 47.5 & 31.0 & 35.2 & 71.3 & 38.9 \\
 & $1\text{k} \times 64$ & 11.2 & 45.0 & 30.8 & 34.7 & 72.8 & 38.9 \\
 & $2\text{k} \times 32$ & 10.8 & 48.6 & 29.5 & 34.6 & 72.9 & 39.3 \\
 & $4\text{k} \times 16$ & 9.6 & 50.2 & 29.6 & 34.0 & 71.8 & 39.0 \\
 & $8\text{k} \times 8$ & 10.4 & 48.6 & 29.8 & 34.5 & 72.3 & 39.1 \\
\multirow{-6}{*}{\textbf{Qwen2.5-7B}} & $16\text{k} \times 4$ & 9.6 & 47.0 & 29.1 & 36.1 & 73.4 & 39.0 \\ \midrule

 & $500 \times 128$ & 3.7 & 33.9 & 27.4 & 22.7 & 57.8 & 29.1 \\
 & $1\text{k} \times 64$ & 5.6 & 31.4 & 27.5 & 23.0 & 58.6 & 29.2 \\
 & $2\text{k} \times 32$ & 5.6 & 35.2 & 26.7 & 22.8 & 55.1 & 29.1 \\
 & $4\text{k} \times 16$ & 4.2 & 34.5 & 26.3 & 22.9 & 57.1 & 29.0 \\
 & $8\text{k} \times 8$ & 4.0 & 34.8 & 26.6 & 22.3 & 56.9 & 28.9 \\
\multirow{-6}{*}{\textbf{Llama3.1-8B-Instruct}} & $16\text{k} \times 4$ & 4.6 & 34.2 & 26.3 & 22.3 & 56.5 & 28.8 \\ \bottomrule

\end{tabular}
}
\end{table}

\section{Limitations}
While this study provides valuable insights into the mechanisms of Long-CoT distillation through the Behavioral Hypothesis, it has a few notable limitations.
First, the experiments were primarily conducted on small-to-medium open-source models, such as the Qwen3 series (ranging from 1.7B to 14B parameters), Qwen2.5-7B, and Llama-3.1-8B-Instruct; it remains to be seen if these behavioral alignment dynamics scale identically to massively larger frontier models.
Although \Cref{tab:teacher_robustness} also includes a Gemma3-12B-it teacher and Qwen3-4B thinking-mode traces, much larger frontier teachers have not been studied.

Second, regarding the taxonomy of factual errors. In our experiments, the ``incorrect'' trajectories were generated by a competent teacher model and filtered by final answer exact-match. As illustrated in \Cref{fig:case_study}, this means our negative set contains many ``near-correct'' traces that exhibit strong logical decomposition but suffer from late-stage arithmetic slips or localized indexing hallucinations. Therefore, our conclusion is most directly applicable to the robustness against localized factual flaws rather than a complete collapse of reasoning. Future work should develop a more granular taxonomy of error types (e.g., arithmetic slips, logically invalid decompositions, contradictory branches, irrelevant rambling) to systematically isolate which specific categories of ``incorrectness'' strictly preserve behavioral alignment and which actively degrade it.



\clearpage
\section*{NeurIPS Paper Checklist}

\begin{enumerate}

\item {\bf Claims}
    \item[] Question: Do the main claims made in the abstract and introduction accurately reflect the paper's contributions and scope?
    \item[] Answer: \answerYes{} 
    \item[] Justification: In the abstract and conclusion of the introduction.

\item {\bf Limitations}
    \item[] Question: Does the paper discuss the limitations of the work performed by the authors?
    \item[] Answer: \answerYes{} 
    \item[] Justification: See Appendix D.

\item {\bf Theory assumptions and proofs}
    \item[] Question: For each theoretical result, does the paper provide the full set of assumptions and a complete (and correct) proof?
    \item[] Answer: \answerNA{} 
    \item[] Justification: N/A

    \item {\bf Experimental result reproducibility}
    \item[] Question: Does the paper fully disclose all the information needed to reproduce the main experimental results of the paper to the extent that it affects the main claims and/or conclusions of the paper (regardless of whether the code and data are provided or not)?
    \item[] Answer: \answerYes{} 
    \item[] Justification: See experimental setups.

\item {\bf Open access to data and code}
    \item[] Question: Does the paper provide open access to the data and code, with sufficient instructions to faithfully reproduce the main experimental results, as described in supplemental material?
    \item[] Answer: \answerNo{}
    \item[] Justification: Code and representative training/evaluation data are publicly available at \url{https://github.com/yuyongcan/Long-CoT-Distillation} and \url{https://huggingface.co/datasets/yuyongcan/Long-CoT-Distillation}. The released configurations cover selected experiments rather than every model and domain in the paper.

\item {\bf Experimental setting/details}
    \item[] Question: Does the paper specify all the training and test details (e.g., data splits, hyperparameters, how they were chosen, type of optimizer) necessary to understand the results?
    \item[] Answer: \answerYes{} 
    \item[] Justification: See experimental setups.

\item {\bf Experiment statistical significance}
    \item[] Question: Does the paper report error bars suitably and correctly defined or other appropriate information about the statistical significance of the experiments?
    \item[] Answer: \answerYes{} 
    \item[] Justification: We report the avg@16 to mitigate the variance.

\item {\bf Experiments compute resources}
    \item[] Question: For each experiment, does the paper provide sufficient information on the computer resources (type of compute workers, memory, time of execution) needed to reproduce the experiments?
    \item[] Answer: \answerYes{} 
    \item[] Justification: We use 8 A100-80G GPUs.
    
\item {\bf Code of ethics}
    \item[] Question: Does the research conducted in the paper conform, in every respect, with the NeurIPS Code of Ethics \url{https://neurips.cc/public/EthicsGuidelines}?
    \item[] Answer: \answerYes{} 
    \item[] Justification: N/A

\item {\bf Broader impacts}
    \item[] Question: Does the paper discuss both potential positive societal impacts and negative societal impacts of the work performed?
    \item[] Answer: \answerNA{} 
    \item[] Justification: N/A
    
\item {\bf Safeguards}
    \item[] Question: Does the paper describe safeguards that have been put in place for responsible release of data or models that have a high risk for misuse (e.g., pre-trained language models, image generators, or scraped datasets)?
    \item[] Answer: \answerNA{} 
    \item[] Justification: N/A

\item {\bf Licenses for existing assets}
    \item[] Question: Are the creators or original owners of assets (e.g., code, data, models), used in the paper, properly credited and are the license and terms of use explicitly mentioned and properly respected?
    \item[] Answer: \answerYes{} 
    \item[] Justification: N/A

\item {\bf New assets}
    \item[] Question: Are new assets introduced in the paper well documented and is the documentation provided alongside the assets?
    \item[] Answer: \answerNA{} 
    \item[] Justification: N/A

\item {\bf Crowdsourcing and research with human subjects}
    \item[] Question: For crowdsourcing experiments and research with human subjects, does the paper include the full text of instructions given to participants and screenshots, if applicable, as well as details about compensation (if any)? 
    \item[] Answer: \answerNA{} 
    \item[] Justification: N/A

\item {\bf Institutional review board (IRB) approvals or equivalent for research with human subjects}
    \item[] Question: Does the paper describe potential risks incurred by study participants, whether such risks were disclosed to the subjects, and whether Institutional Review Board (IRB) approvals (or an equivalent approval/review based on the requirements of your country or institution) were obtained?
    \item[] Answer: \answerNA{} 
    \item[] Justification: N/A

\item {\bf Declaration of LLM usage}
    \item[] Question: Does the paper describe the usage of LLMs if it is an important, original, or non-standard component of the core methods in this research? Note that if the LLM is used only for writing, editing, or formatting purposes and does \emph{not} impact the core methodology, scientific rigor, or originality of the research, declaration is not required.
    \item[] Answer: \answerNA{} 
    \item[] Justification: N/A

\end{enumerate}

\end{document}